\documentclass[12pt,onecolumn,twoside]{IEEEtran}
\usepackage{psfig}
\usepackage{floatflt} 
\usepackage{paralist} 
\usepackage{latexsym}
\usepackage{amsfonts}
\usepackage{amssymb}
\usepackage{amstext}

\newcommand{\bA}{{\mathbf{A}}}
\newcommand{\bB}{{\mathbf{B}}}

\newcommand{\bM}{{\mathbf{M}}}
\newcommand{\bN}{{\mathbf{N}}}

\newcommand{\bR}{{\mathbf{R}}}
\newcommand{\bS}{{\mathbf{S}}}

\newcommand{\bU}{{\mathbf{U}}}
\newcommand{\bV}{{\mathbf{V}}}
\newcommand{\bW}{{\mathbf{W}}}

\newcommand{\ba}{{\mathbf{a}}}
\newcommand{\bb}{{\mathbf{b}}}

\newcommand{\bm}{{\mathbf{m}}}

\newcommand{\bt}{{\mathbf{t}}}

\newcommand{\bw}{{\mathbf{w}}}

\newcommand{\bSigma}    {{\mathbf{\Sigma}}}

\newcommand{\bone}      {{\mathbf{1}}}

\newcommand{\cR}{{\mathcal{R}}}
\newcommand{\cS}{{\mathcal{S}}}

\begin{document}

\title{3-D Rigid Models from Partial Views---\\---Global Factorization}

\author{Pedro M.~Q.~Aguiar,~{\sl Member,~IEEE,} \ \ Rui~F.~C.~Guerreiro, \\ and Bruno~B.~Gon\c{c}alves
\thanks{Contact author: P. Aguiar, ISR---Institute for Systems and Robotics, Instituto
Superior T\'{e}cnico, Av. Rovisco Pais, 1049-001 Lisboa, Portugal.
E-mail: {\tt aguiar@isr.ist.utl.pt}. His work was partially
supported by FCT grant POSI/SRI/41561/2001 and FCT
ISR/IST plurianual funding, POSC, FEDER.}
\thanks{R. Guerreiro is
with Philips Research, Eindhoven.}
\thanks{B. Gon\c{c}alves is with WhatEverSoft, WhatEverNet.}
}


\date{January, 2005}

\maketitle

\begin{abstract}
The so-called factorization methods recover 3-D~rigid structure
from motion by factorizing an observation matrix that collects
2-D~projections of features. These methods became popular due to
their robustness---they use a large number of views, which
constrains adequately the solution---and computational
simplicity---the large number of unknowns is computed through an
SVD, avoiding non-linear optimization. However, they require that
all the entries of the observation matrix are known. This is
unlikely to happen in practice, due to self-occlusion and limited
field of view. Also, when processing long videos, regions that
become occluded often appear again later. Current factorization
methods process these as new regions, leading to less accurate
estimates of 3-D structure. In this paper, we propose a global
factorization method that infers \underline{complete 3-D models}
directly from the 2-D projections in the \underline{entire set of
available video frames}. Our method decides whether a region that
has become visible is a region that was seen before, or a
previously unseen region, in a global way, {\it i.e.}, by seeking
the \underline{simplest rigid object} that describes well the
entire set of observations. This global approach increases
significantly the accuracy of the estimates of the 3-D shape of
the scene and the 3-D motion of the camera. Experiments with
artificial and real videos illustrate the good performance of our
method.
\end{abstract}






\section{Introduction}
\label{sec:int}

In areas ranging from virtual reality and digital video to
robotics, an increasing number of applications need
three-dimensional (3-D) models of real-world objects. Although
expensive active sensors, {\it e.g.}, laser rangefinders, have
been frequently used to acquire 3-D~data, in many relevant
situations only two-dimensional (2-D) video data is available and
the 3-D models have to be recovered from their 2-D~projections. In
this paper, we address the automatic recovery of complete
3-D~models from 2-D video sequences.

\subsection{Related work and motivation}
Since the strongest cue to infer 3-D shape from video is the 2-D
motion of the image brightness pattern, our problem has been often
referred to as structure from motion (SFM). Since using a large
number of views, rather than simply two consecutive frames, leads
to more constrained problems, thus to more accurate 3-D models,
current research has been focused on multi-frame~SFM. The
factorization method introduced in~\cite{tomasi92} overcomes the
difficulties of multi-frame SFM---nonlinearity and large number of
unknowns---by using matrix subspace projections.
In~\cite{tomasi92}, the trajectories of feature points are
collected into an observation matrix that, due to the rigidity of
the 3-D object, is highly rank deficient in a noiseless situation.
The 3-D shape of the object and the 3-D motion of the camera are
recovered from the rank deficient matrix that best matches the
observation matrix. The work of~\cite{tomasi92} was extended in
several ways, {\it e.g.}, geometric projection
models~\cite{sturm96,poelman97}, 3-D shape
primitives~\cite{shapiro95,quan96,aguiar01}, recursive
formulation~\cite{morita97},
uncertainty~\cite{morris98,aguiar99-1,irani02,aguiar03}, multibody
scenario~\cite{costeira98}. Besides recovering 3-D~rigid structure
from video sequences, recent approaches to several other image
processing/computer vision problems require determining linear or
affine low dimensional subspaces from noisy observations. These
problems include object recognition~\cite{basri88,ullman91},
applications in photometry~\cite{shashua92,moses93,belhumeur96},
and image alignment~\cite{manor99-1,manor99-2}.

In general, such low dimensional subspaces are found by estimating
rank deficient matrices from noisy observations of their entries.
When the observation matrix is completely known, the solution to
this problem is easily obtained from its {\it Singular Value
Decomposition}~(SVD)~\cite{golub96}. However, in practice, the
observation matrix may be incomplete, {\it i.e.}, some of its
entries may be unknown (unobserved). When recovering 3-D structure
from video, the observation matrix collects 2-D trajectories of
projections of feature
points~\cite{tomasi92,sturm96,poelman97,morita97,irani02,aguiar03}
or other primitives~\cite{shapiro95,quan96,morris98,aguiar01}. In
real life video clips, these projections are not visible along the
entire image sequence due to the occlusion and the limited field
of view. Thus, the observation matrix is in general incomplete.
For the specific cases we focus in this paper---videos that show
views all around (non-transparent) 3-D objects---this incomplete
characteristic of the observation matrix is particularly
noticeable.

The problem of estimating a rank deficient matrix from noisy
observations of a subset of its entries deserved then the
attention of several researchers. References~\cite{tomasi92}
and~\cite{jacobs97} propose sub-optimal solutions.
In~\cite{tomasi92}, the missing values of the observation matrix
are ``filled in", in a sequential way, by using SVDs of observed
submatrices. In~\cite{jacobs97}, the author proposes a method to
combine the constrains that arise from the observed submatrices of
the original matrix. More recently, several bidirectional
optimization schemes were proposed, {\it e.g.},
\cite{maruyama99,guerreiro02,Brandt:smvp2002,Aanaes:etal:pami2002,Huynh:etal:iccv2003,Vidal:Hartley:cvpr2004,chen04},
and the problem was also addressed by using nonlinear
optimization~\cite{buchanan05}.

However, none of the methods above deal with ``self-inclusion",
{\it i.e.}, with the fact that a region that disappears due to
self-occlusion may appear again later. When this happens, the
re-appearing region is usually treated as a new region, {\it
i.e.}, as a region that was never seen before. This procedure has
two drawbacks: {\bf i)}~the problem becomes less constrained than
it should, thus leading to less accurate estimates of 3-D
structure; and {\bf ii)} further 3-D processing is needed to fuse
the recovered multiple versions of the same real-world regions.

%
%
%
%

\subsection{Proposed approach}
We propose a global approach to recover complete 3-D models from
video. Global in the sense that our method computes the {\it
simplest} 3-D rigid object that best matches the {\it entire} set
of 2-D image projections. This way we avoid having to post-process
several partial 3-D models, each obtained from a smaller set of
frames, or an inaccurate 3-D model obtained from the entire set of
frames without detecting re-appearing regions. We develop a global
cost function that balances two terms---model fidelity and
complexity penalization. The model fidelity term measures the
error between the model---a 3-D shape and a set of re-appearing
regions---and the observations, as in Maximum Likelihood~(ML)
estimation. This error is simply given by the distance of a
re-arranged observation matrix to the appropriate space of rank
deficient matrices. The penalty term measures the complexity of
the 3-D model, which is easily coded by the number of feature
points used to describe the observations. By minimizing this
global cost, we get what statisticians usually call a Penalized
Likelihood~(PL)~\cite{green98penalized} estimate of the 3-D
structure. Through PL estimation, re-appearing regions are then
detected when the increase of the complexity of the 3-D model does
not compensate a slightly better fit to the observations, meaning
that a more complex 3-D model would fit the observation noise
rather than the 3-D real-world object.

The minimization of the PL cost function requires the computation
of rank deficient matrices from partial observations. In the
paper, we study two iterative algorithms developed for this
problem. The first algorithm is based in a well known method to
deal with missing data---the {\it
Expectation-Maximization}~(EM)~\cite{KN:McLachlanKrishnan}.
Although the authors don't refer it, the bidirectional scheme
in~\cite{maruyama99} is also EM-based. However, as detailed below,
the EM algorithm we propose is more general and computationally
simpler than the one in~\cite{maruyama99}. Our second two-step
iterative scheme is similar to Wiberg's algorithm~\cite{wiberg76}
and related to the one used in~\cite{shum95} to model polyhedral
objects. It computes, alternately, in closed form, the row space
matrix and the column space matrix whose product is the solution
matrix. We call this the {\it Row-Column}~(RC) algorithm.

In the paper, we illustrate the behavior of both iterative
algorithms with simple cases and evaluate their performance with
more extensive experiments. In particular, we study the impact of
the initialization on the algorithm's behavior. From these
experiments, we conclude that the RC algorithm is more robust than
EM in what respects to the sensibility to the initialization.
Furthermore, the number of iterations needed for good convergence
and the computational cost of each iteration are both smaller for
RC than for EM. Obviously, the performances of both EM and RC
improve when the initial estimate provided to the algorithms is
more accurate. Any sub-optimal method, {\it e.g.} the ones
in~\cite{tomasi92,jacobs97}, can be used to compute such an
initialization. Our experience shows that, with a simple
initialization procedure, even for high levels of noise and large
amount of missing data, both iterative algorithms converge: {\bf
i)}~to the global optimum; and {\bf ii)}~in a very small number of
iterations. Naturally, under our global factorization approach,
the rank deficient matrices can also be computed by using
nonlinear methods such as~\cite{buchanan05}.

We use EM and RC in the context of recovering complete 3-D~models
from video, through the minimization of our global PL cost. Our
experiments show that fitting the rank deficient matrix to the
entire (re-arranged) observation matrix (which, naturally, misses
several entries) leads to better 3-D~reconstructions than those
obtained by combining partial 3-D models estimated by fitting
submatrices to smaller subsets of data (each corresponding to a
subset of features that were visible in a subset of consecutive
frames).

\subsection{Paper organization} In section~\ref{sec:sfm} we
overview the matrix factorization approach to the recovery of 3-D
structure. Section~\ref{sec:c3-Dm} details the problem addressed
in this paper---the direct recovery of complete \linebreak {3-D}
models from incomplete 2-D~trajectories of features points. In
section~\ref{sec:pl} we describe our solution as the minimization
of a global cost function. Section~\ref{sec:missing} describes the
algorithms used to estimate rank deficient matrices from
incomplete observations. In sections~\ref{sec:exp1}
and~\ref{sec:exp2}, we present experiments and an application to
video compression. Section~\ref{sec:conc} concludes the paper.

Preliminary versions of parts of this work appeared
in~\cite{guerreiro02,guerreiro02a,guerreiro03,goncalves04}.

\section{Matrix Factorization for 3-D Structure from Motion}\label{sec:sfm}

In the original factorization method~\cite{tomasi92}, the authors
track $P$~feature points over $F$~frames and collect their 2-D
trajectories in the $2F\times P$~observation matrix~$\bW$. The
observation matrix is written in terms of the parameters that
describe the 3-D~structure (3-D~shape and 3-D motion)~as
\begin{equation}\label{eq:w}
\bW=\bR\bS^T+\bt\bone=\left[\begin{array}{c|c} \bR & \bt
\end{array}\right]\left[\begin{array}{c} \bS^T
\\\hline
\bone\end{array}\right]+\mbox{noise}\,,
\end{equation}
where~$\bR$ and~$\bt$ represent the rotational and translational
components of the 3-D~motion of the camera and~$\bS$ represents
the 3-D~shape of the scene. Matrix~$\bR$ is~$2F\times 3$. It
collects entries of the $2F$ 3-D~rotation matrices that code the
camera orientations. The $2F\times 1$ vector~$\bt$ collects the
$2F$~camera positions. Matrix~$\bS$ is~$P\times 3$. It contains
the 3-D~coordinates of the $P$~feature points. See~\cite{tomasi92}
for the details.

The problem of recovering 3-D~structure from motion~(SFM) is then:
given~$\bW$, estimate~$\bR$, $\bS$, and~$\bt$. Due to the specific
structure of the entries of the 3-D rotation matrices in~$\bR$,
see for example~\cite{ayache91}, this is an highly non-linear
problem. The factorization method uses a subspace approximation to
derive an efficient solution to this problem. In~\cite{tomasi92},
the authors noted that,
in a noiseless situation, the $2F\times P$ observation
matrix~$\bW$ in~(\ref{eq:w}) belongs to a restricted
subspace---the space of the rank~4 matrices. In practice, due to
the noise, although the observations matrix is in general full
rank, it is well approximated by a rank~4 matrix. The method
in~\cite{tomasi92} recovers SFM performing two separate steps. The
first step computes the rank~4 matrix that best matches the
observation matrix~$\bW$. The second step performs a normalization
that approximates the constraints imposed by the entries of the
rotation matrices in~$\bR$.

To compute the rank~4 matrix that best matches the observations,
most common techniques use the Singular Value Decomposition~(SVD)
of~$\bW$. This requires that all entries of~$\bW$ are known, which
means that the projections of the feature points to be processed
were observed in all frames, {\it i.e.}, that they are seen during
the entire video sequence. In real world applications, this
assumption limits severely the feature point candidates because
very often important regions that are seen in some frames are not
seen in others due to the scene self-occlusion and the limited
field of view.

\section{Recovering Complete 3-D Models: Occlusion and Re-appearing Features}
\label{sec:c3-Dm}

Since in many practical situations, several regions of the 3-D
scene to reconstruct do not appear in the entire set of images of
the video sequence, the observation matrix has several unknown
entries. Naturally, a suboptimal solution to the recovery of 3-D
structure in such situations is to process separately several
completely known submatrices of the observations matrix. However,
this leads to a less constrained problem---the global rigidity of
the scene is not appropriately modelled.

Another aspect that must be taken into account is the fact that
regions that disappear may re-appear later. For example, when
attempting to build a complete model of a 3-D object, we must use
a video stream containing views that completely ``cover" the
object, typically, a video obtained by rotating the camera around
the object. Obviously, as the camera moves, some feature points
disappear, due to object self-occlusion, remain invisible during
certain period and then re-appear. In general, each feature point
has thus several tracking periods. Although this is always the
case when constructing complete 3-D models, it also happens very
frequently when processing real-life videos in general.

\begin{figure}[hbt]
\centerline{\psfig{figure=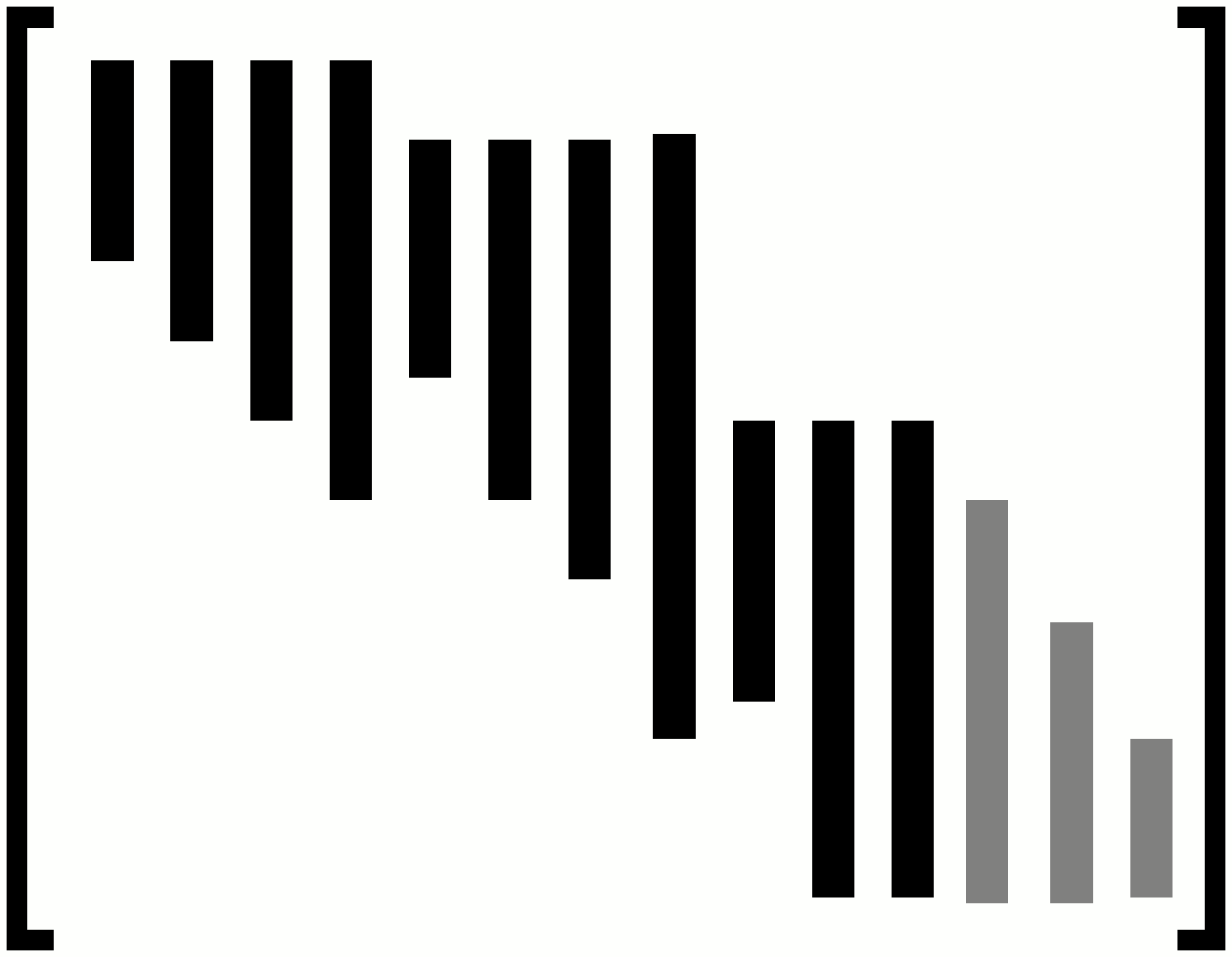,height=4cm}\hspace*{2cm}
\psfig{figure=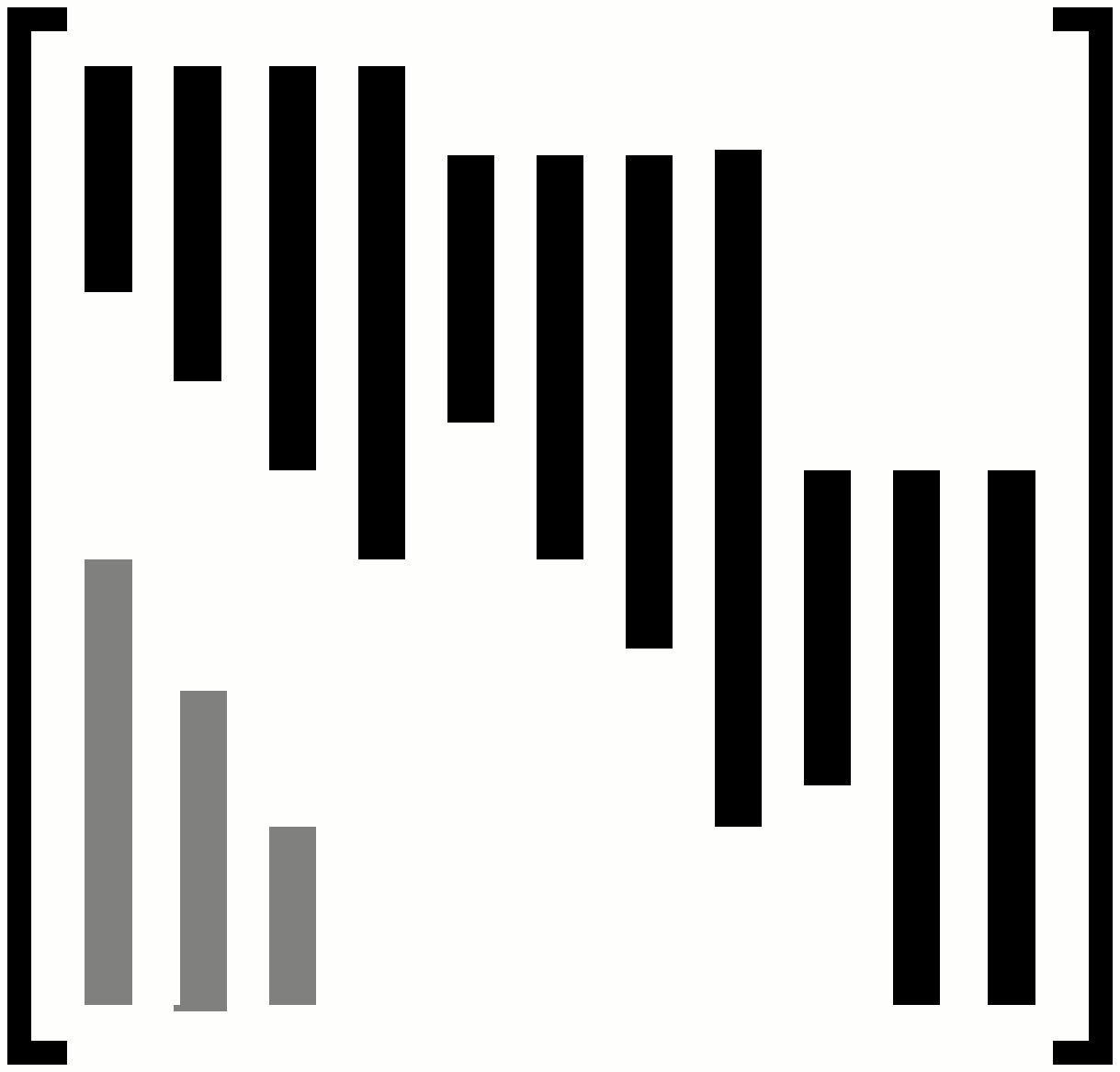,height=4cm}}
\caption{Left: original observation matrix $\bW$.
Right: re-arranged observation matrix $\bW_R$, after detecting
re-appearing features.\label{fig:matrix}}
\end{figure}

\begin{figure}[hbt]
\centerline{\psfig{figure=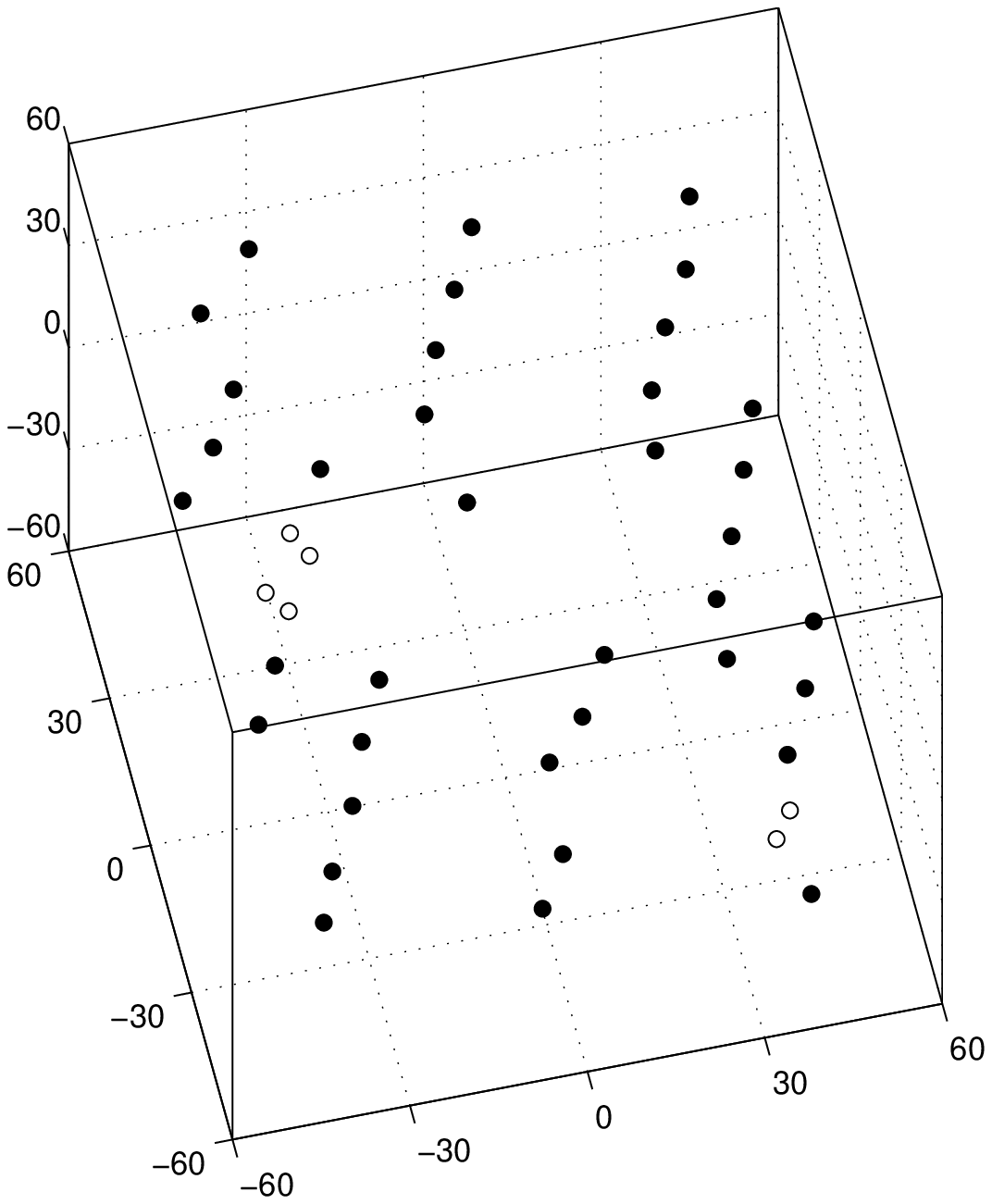,width=6cm}\hspace*{2cm}
\psfig{figure=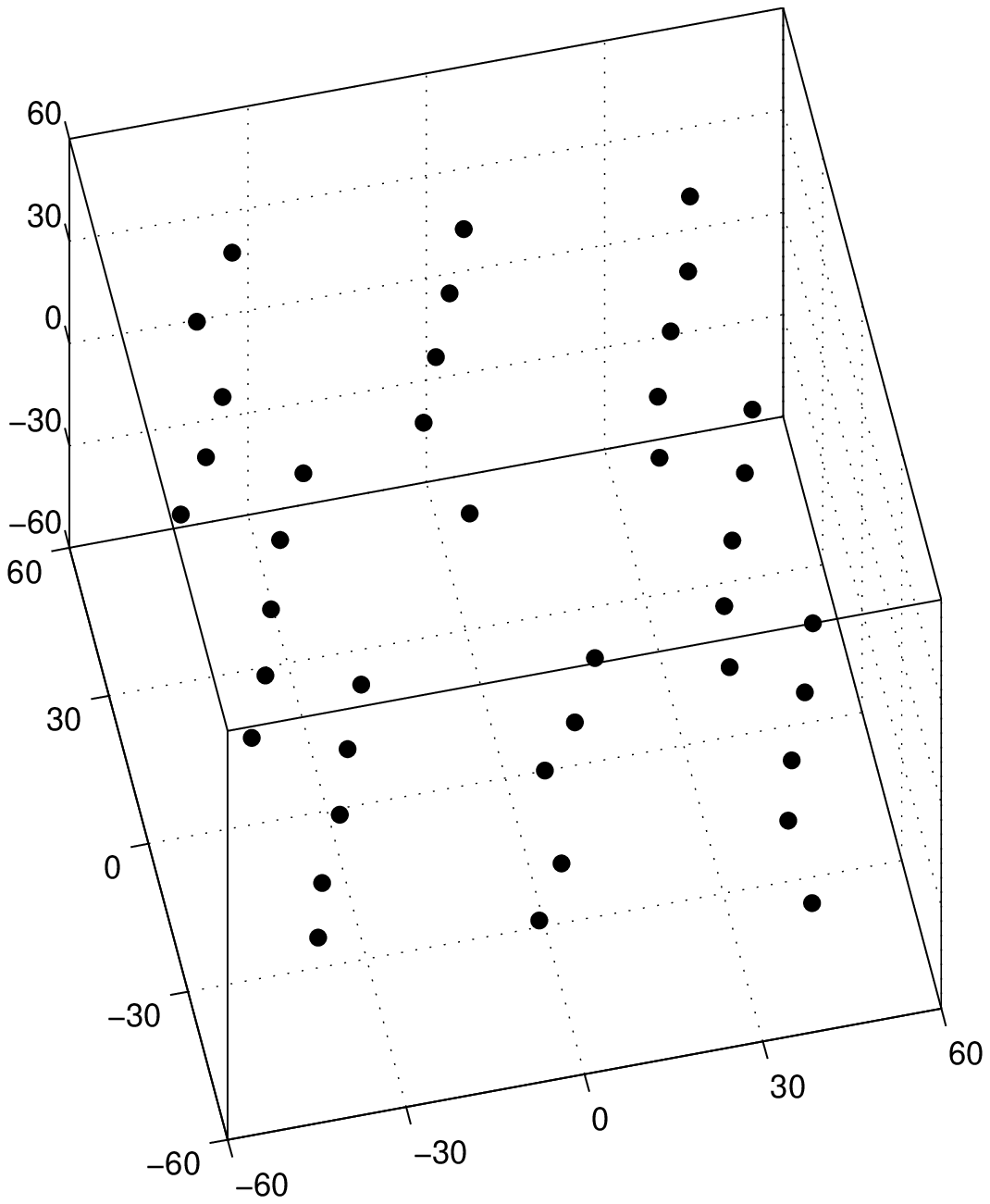,width=6cm}} 
\caption{3-D shape recovered from the original matrix~$\bW$, {\it
i.e.}, without detecting re-appearing features (left) and from the
re-arranged matrix~$\bW_R$, {\it i.e.}, after detecting
re-appearing features (right). In the left plot, three feature
points are erroneously recovered as 3-D point pairs (marked with
small circles). See in Fig.~\ref{fig:matrix} the representation of
the known entries of~$\bW$ and~$\bW_R$.
\label{fig:3-Dshape_wrong}}
\end{figure}

Even current methods that deal with occlusion, {\it e.g.},
\cite{jacobs97,maruyama99,guerreiro02,Brandt:smvp2002,Aanaes:etal:pami2002,Huynh:etal:iccv2003,Vidal:Hartley:cvpr2004,chen04,buchanan05},
do not consider a re-appearing feature as another observation of a
previously seen point. They rather consider as many
%
%
%
%
%
feature points as tracking periods. To illustrate this point, we
represent on the left image of Fig.~\ref{fig:matrix} the typical
shape of the known entries of the observation matrix~$\bW$. Each
feature trajectory is represented by a column of~$\bW$. The three
last columns (in gray) correspond to re-appearing features, {\it
i.e.}, they are second tracking periods of the features that were
first tracked and collected in the three first columns.
When a given feature has two tracking periods, current methods
like the ones in the references above, return a 3-D shape
containing two 3-D feature points that correspond to the same 3-D
point of the real-world object. Although these two 3-D feature
points would coincide in a noiseless situation, in practice they
are just close to each other. To demonstrate this, we represent on
the left plot of Fig.~\ref{fig:3-Dshape_wrong}, the 3-D shape
recovered from a set of synthesized trajectories of 40 features,
by using the method in~\cite{guerreiro02}. To simulate occlusion,
three of the trajectories were artificially ``interrupted",
leading to three pairs of recovered 3-D points, marked with small
circles in the left plot of Fig.~\ref{fig:3-Dshape_wrong}.


%

%
%
%

A simple way to detect re-appearing features is based on a local
analysis of the distance between recovered 3-D points. However,
this procedure is hard in practice due to the sensitivity to the
threshold below which the features would be considered to
correspond to the same 3-D point. In fact, the distance between
the features that correspond to the same 3-D point, depends not
only on the noise level but also on the camera-object distance,
which is very difficult to estimate accurately enough.
%
%
%
The detection of re-appearing features must then be based on a
global approach.

Our approach in this paper is then to re-arrange the observation
matrix~$\bW$ into a smaller matrix~$\bW_R$ that merges all the
tracking periods of the same feature in the same column. For the
observation matrix~$\bW$ shown on the left side of
Fig.~\ref{fig:matrix}, the re-arranged matrix~$\bW_R$ would be as
shown on its right. Finding matrix~$\bW_{R}$ is equivalent to
detecting the re-appearing features. Note that the statements of
section~\ref{sec:sfm} remain valid for the re-arranged
matrix~$\bW_R$; in particular,
 $\bW_{R}$ is rank 4 in a noiseless situation, just
like the original observation matrix~$\bW$. As pointed out before,
the advantage of using~$\bW_{R}$ rather than~$\bW$ is that the SFM
problem becomes more constrained, thus leading to more accurate
estimates of the 3-D~structure.

\section{Global Approach: Penalized Likelihood Estimation}\label{sec:pl}

We formulate the detection of re-appearing features as a model
selection problem, where a model is represented by a re-arranged
observation matrix~$\bW_{r}$. Matrix~$\bW_{r}$ determines the
number of feature points of the 3-D~model, $P_{r}$, and the
correspondences between columns of the original observation
matrix~$\bW$ and points of the 3-D real-world object. To select
the best model~$\bW_{R}$, we propose a global approach---we seek
the simplest 3-D rigid object that describes the entire set of
(incomplete) feature trajectories.

%
%

\subsection{Penalized likelihood cost}
In our approach, re-appearing features are detected through the
minimization of a global cost-function that takes into account
both the accuracy of the model and its complexity. Thus, the
re-arranged matrix~$\bW_R$ is given~by
\begin{equation}
\bW_R=\arg\min_{\bW_r\in\cR(\bW)}\biggl\{E_{\cS_4}\left(\bW_r\right)+\alpha
P_{r}\biggr\},\label{eq:cost}
\end{equation}
where~$E_{\cS_4}\left(\bW_r\right)$ measures the error of the
model~$\bW_r$ as its distance to the space of rank~4 matrices,
{\it i.e.}, it is a likelihood term, and~$P_r$ is the number of
feature points of the model, {\it i.e.}, it codes the model
complexity (matrix $\bW_r$ is $2F\times P_r$). The parameter
$\alpha$ balances the two terms. Naturally, the minimization
in~(\ref{eq:cost}) is constrained to the set of matrices that can
be found by re-arranging the original observations matrix~$\bW$.
We denote this set by~$\cR(\bW)$.

The likelihood term~$E_{\cS_4}\left(\bW_r\right)$
in~(\ref{eq:cost}) models the object rigidity. As outlined in
section~\ref{sec:sfm}, in a noiseless situation, the matrix
collecting the trajectories of points of a rigid object is rank~4.
This term is then naturally given by the distance of the matrix
$\bW_r$ to the space of the $2F\times P_r$ rank~4 matrices. Since
not all the entries of $\bW_r$ are known, we measure this distance
by masking the unknown entries:
\begin{equation}
E_{\cS_4}(\bW_r)=\min_{\widetilde{\bW}\in
\cS_4}\left\|\left(\bW_r-\widetilde{\bW}\right)\odot\bM_r\right\|_F,\label{eq:l}
\end{equation}
where~$\left\|.\right\|_F$ represents the Frobenius norm, which
arises from assuming that the observations are corrupted by white
Gaussian noise, $\cS_4$~denotes the space of the $2F\times P_r$
rank~$4$ matrices, $\odot$~represents the elementwise product,
also known as the Hadamard product, and the matrix~$\bM_r$ is a
binary mask that accounts for the known entries of the
matrix~$\bW_r$, {\it i.e.}, the entry $(i,j)$ of~$\bM_r$ is
$m_{ij}=1$ if the entry~$w_{ij}$ of~$\bW_r$ is known and
$m_{ij}=0$ otherwise.

In Bayesian inference approaches, the term~$\alpha P_{r}$
in~(\ref{eq:cost}) is considered to be a prior and the
minimization~(\ref{eq:cost}) leads to the Maximum a Posteriori
(MAP) estimate~\cite{berger93} (the cost function
in~(\ref{eq:cost}) is the negative log posterior probability).
Penalization terms, such as~$\alpha P_{r}$ in~(\ref{eq:cost}),
have also been introduced in the model selection literature
through information-theoretic criteria like
Akaike's~AIC~\cite{ripley96} or the Minimum Description Length
(MDL) principle~\cite{barron98}. Naturally, different basic
principles lead to different choices for the balancing
parameter~$\alpha$ but the structure of the cost function is most
often as in~(\ref{eq:cost}). This generic form~(\ref{eq:cost}) is
usually called by the statisticians a {\it penalized
likelihood}~(PL) cost function~\cite{green98penalized}. In our
case, to find a valid range for the weight parameter~$\alpha$, we
performed several experiences. By testing with pertinent values
for the number of features, number of images, noise level, $\%$ of
missing data, and number of re-appearing features, we concluded
that there is a wide range of values for $\alpha$ that
%
%
leads to a good balance between maximizing the number of correct
detections of re-appearing features (probability of detection) and
minimizing the number of incorrect detections (probability of
false alarm).

\subsection{Minimization algorithm}

To find candidates $\bW_r\in\cR(\bW)$, our algorithm starts by
selecting from the original observation matrix~$\bW$, pairs of
columns that can be merged with each other. Naturally, these pairs
correspond to disjoint tracking periods. For example, for the
matrix~$\bW$ in the right side of Fig.~\ref{fig:matrix}, each one
of the six last columns could be merged with the first column,
therefore, in what respects to the feature corresponding to the
first column, there are seven possible situations: it could either
have re-appeared, generating one of the trajectories of the six
last columns, or remain occluded for the remaining of the video.

To obtain a computationally feasible algorithm, we prune the
search---our algorithm decides by
%
%
%
comparing the costs of merging each selected pair of columns with
the one of considering that there are not re-appearing features,
{\it i.e.}, of the model~$\bW_{r} = \bW$. The process is then
repeated until the cost~(\ref{eq:cost}) does not decrease by
merging any selected pair of columns. The search could be further
pruned, {\it e.g.}, using the local approach outlined in the
previous section to guide the process, thus testing only pairs of
columns corresponding to feature points that are close in 3-D.


What remains to be addressed is the computation of the likelihood
term~$E_{\cS_4}(\bW_r)$, given by expression~(\ref{eq:l}). The
minimization in~(\ref{eq:l}) requires finding the rank~4 matrix
that best matches the incompletely known matrix~$\bW_r$. There is
not known closed-form solution for this problem. In the following
section, we develop iterative algorithms that compute such rank
deficient approximations in very few iterations. The likelihood
term~$E_{\cS_4}(\bW_r)$ is then computed~as
\begin{equation}
E_{\cS_4}(\bW_r)=\left\|\left(\bW_r-\widehat{\bW_r}\right)\odot\bM_r\right\|_F,
\label{eq:ll}
\end{equation}
where $\widehat{\bW_r}$ is the rank~4 matrix that best matches the
known entries of the matrix $\bW_r$, {\it i.e.},
\begin{equation}
\widehat{\bW_r}=\arg\min_{\widetilde{\bW}\in
\cS_4}\left\|\left(\bW_r-\widetilde{\bW}\right)\odot\bM_r\right\|_F
. \label{eq:approx}
\end{equation}
The iterative algorithms we describe in the sequel converge in
very few iterations to $\widehat{\bW_r}$, when adequately
initialized. In the appendix, we describe a strategy to compute
such initialization. We reduce the computational burden of the
global approach by performing this initialization step only once,
when testing the model that corresponds to the original
observation matrix, $\bW_{r} = \bW$, and using the same initial
guess when testing the other models.


\section{Estimation of rank deficient matrices with Missing Data}
\label{sec:missing}

We now address the estimation of the rank deficient
matrix~$\widehat{\bW_r}$ given by expression~(\ref{eq:approx}). To
gain insight and introduce notation, let us  first consider the
case where the $2F\times P_r$~matrix~$\bW_r$ is completely known,
{\it i.e.}, $m_{ij}=1,\forall_{i,j}\Leftrightarrow
\bM_r=\bone_{2F\times P_r}$. In this case, (\ref{eq:approx}) is
simplified to
%
%
\begin{equation} \label{eq:minsvd}
\widehat{\bW_r}=\arg\min_{\widetilde{\bW}\in \cS_4
}\left\|\bW_r-\widetilde{\bW}\right\|_F .
\end{equation}
%
The solution~$\widehat{\bW_r}$ of~(\ref{eq:minsvd}) is known---it
is obtained by selecting the $4$~largest singular values, and the
associated singular vectors, from the Singular Value
Decomposition~(SVD), $\bW_r\!=\!\bU\bSigma\bV$, of the $2F\times
P_r$~matrix~$\bW_r$~\cite{golub96}. We denote this optimal rank
reduction, {\em i.e.}, the projection onto~$\cS_4$,
by~$\bW_r\!\downarrow\!\cS_4$:
\begin{equation}  \label{eq:svd}
\widehat{\bW_r}=\bW_r\!\downarrow\!\cS_4=\bU_{2F\times
4}\,\bSigma_{4\times 4}\,\bV_{4\times P_r} .
\end{equation}

When matrix~$\bW_r$ misses a subset of its entries, the estimation
of~$\widehat{\bW_r}$ leads to the minimization
of~(\ref{eq:approx}), a generalized version of~(\ref{eq:minsvd}).
The existence of unknown entries in~$\bW_r$ prevents us to use the
SVD of~$\bW_r$ as in~(\ref{eq:svd}). In the following subsections
we introduce two iterative algorithms that
minimize~(\ref{eq:approx}). Both algorithms are initialized with a
guess~$\widehat{\bW_r}^{(0)}$, computed as described in the
appendix.

\subsection{Expectation-Maximization algorithm}\label{sec:em}

The {\it Expectation-Maximization} (EM) approach to estimation
problems with missing data works by enlarging the set of
parameters to estimate---the data that is missing is jointly
estimated with the other parameters. The joint estimation is
performed, iteratively, in two alternate steps: {\bf i)}~the {\it
E-step} estimates the missing data given the previous estimate of
the other parameters; {\bf ii)}~the {\it M-step} estimates the
other parameters given the previous estimate of the missing data,
see~\cite{KN:McLachlanKrishnan} for a review on the EM algorithm.

In our case, given the initial estimate~$\widehat{\bW_r}^{(0)}$,
the EM~algorithm estimates in alternate steps: {\bf i)} the
missing entries of the input matrix~$\bW_r$; {\bf ii)} the
rank~$4$ matrix $\widehat{\bW_r}$ that best matches the
``complete" data. The algorithm performs these two steps until
convergence, {\em i.e.}, until the error measured by the Frobenius
norm in~(\ref{eq:approx}) stabilizes.

\subsubsection{E-step---estimation of the missing data}
Given the previous estimate~$\widehat{\bW_r}^{(k-1)}$, the
estimates of the missing
entries~$\left\{{w}_{ij}:m_{ij}\!=\!0\right\}$ of~$\bW_r$ are
simply the corresponding entries~$\widehat{w}_{ij}$
of~$\widehat{\bW_r}^{(k-1)}$. We then build a complete observation
matrix~$\overline{\bW_r}^{(k)}$, whose entry~$\overline{w}_{ij}$
equals the corresponding entry~${w}_{ij}$ of the matrix~$\bW_r$
if~${w}_{ij}$ was observed or its estimate~$\widehat{w}_{ij}$
if~${w}_{ij}$ is unknown,
\begin{equation}\label{eq:mask}
  \overline{w}_{ij}=\left\{\begin{array}{lll}
    {w}_{ij} & \mbox{if}&m_{ij}=1  \\
    \widehat{w}_{ij} & \mbox{if}& m_{ij}=0 ,
\end{array}\right.
\end{equation}
or, in matrix notation,
\begin{equation}\label{eq:EM_eq1}
  \overline{\bW_r}^{(k)}=\bW_r\odot \bM_r + \widehat{\bW_r}^{(k-1)}\odot
  \biggl[\bone-\bM_r\biggr].
\end{equation}

\subsubsection{M-step---estimation of the rank deficient matrix}
We are now given the complete data
matrix~$\overline{\bW_r}^{(k)}$, with the estimates of the missing
data from the E-step. The rank~$4$ matrix~$\widehat{\bW_r}^{(k)}$
that best matches~$\overline{\bW_r}^{(k)}$ in the Frobenius norm
sense, is then obtained from the SVD of~$\overline{\bW}_{k}$, as
in~(\ref{eq:svd}),
\begin{equation}
\widehat{\bW_r}^{(k)}=\overline{\bW_r}^{(k)}\!\downarrow\!\cS_4.
\end{equation}


In reference~\cite{maruyama99}, the authors develop a
bidirectional algorithm to factor out an observation matrix with
missing data, in the context of recovering rigid~SFM. Their
bidirectional algorithm is in fact an EM algorithm developed to
the specific strategy of treating the 3-D~translation separately.
In opposition, the EM~algorithm just described is general, i.e, it
solves any rank deficient matrix approximation problem with
missing data. Furthermore, our E-step in~(\ref{eq:EM_eq1}) is
simpler than the corresponding step of~\cite{maruyama99} that
requires inverting matrices.

\subsection{Row-Column Algorithm}\label{sec:rc}

We now describe the {\it Row-Column} (RC) algorithm---another
iterative approach, similar to Wiberg's algorithm~\cite{wiberg76},
to the estimation of a rank deficient matrix that best matches an
incomplete observation. From our experience, summarized in
section~\ref{sec:exp1}, the RC~algorithm is not only
computationally cheaper than EM, avoiding SVD computations and
exhibiting faster convergence, but also more robust than EM to
initializations far from the solution.

For the RC~algorithm, we parameterize the $2F\times P_r$  rank~$4$
matrix~$\widetilde{\bW}$ in~(\ref{eq:approx}) as the product
\begin{equation}\label{eq:ab}
  \widetilde{\bW}=
  \widetilde{\bA}_{2F\times 4}\,\widetilde{\bB}_{4\times P_r}\in\cS_4,
\end{equation}
where~$\widetilde{\bA}$ determines the column space
of~$\widetilde{\bW}$ and~$\widetilde{\bB}$ its row space. The
rank~4 matrix~$\widehat{\bW_r}$ that best matches~$\bW_r$ is
obtained by minimizing the cost function in~(\ref{eq:approx}) with
respect to~(wrt) the column space and row space matrices, {\em
i.e.},
\begin{equation}  \label{eq:minab}
\widehat{\bW_r}\!=\!\bA\bB,
\qquad\mbox{where}\qquad\left\{\bA,\bB\right\}=
\arg\min_{\widetilde{\bA},\widetilde{\bB}}
\left\|\left(\bW_r-\widetilde{\bA}\widetilde{\bB}\right)\odot\bM_r\right\|_F.
\end{equation}
By using the re-parameterization~(\ref{eq:ab}), we map the
constrained minimization~(\ref{eq:approx}) wrt
$\widetilde{\bW}\!\in\!\cS_4$ into the unconstrained
minimization~(\ref{eq:minab}) wrt~$\widetilde{\bA}$
and~$\widetilde{\bB}$.

We minimize~(\ref{eq:minab}) in two alternate steps: {\bf i)} the
{\it R-step} assumes the column space matrix~$\bA$ is known and
estimates the row space matrix~$\bB$; {\bf ii)} the {\it C-step}
estimates~$\bB$ for known~$\bA$. The algorithm is initialized by
computing~$\bA^{(0)}$ from the initial
estimate~$\widehat{\bW_r}^{(0)}$ and it runs until the value of
the norm in~(\ref{eq:minab}) stabilizes.

When there is no missing data, {\em i.e.}, when~$\bM_r\!=\!\bone$
in~(\ref{eq:minab}), the R- and C- steps above are linear least
squares~(LS) problems whose solutions are obtained by using the
Moore-Penrose pseudoinverse~\cite{golub96},
\begin{equation}\label{eq:pseudob}
\bB^{(k)}=\left({\bA^{(k-1)}}^T{\bA^{(k-1)}}\right)^{-1}{\bA^{(k-1)}}^T\bW_r,\qquad
\bA^{(k)}=\bW_r{\bB^{(k)}}^T\left(\bB^{(k)}{\bB^{(k)}}^T\right)^{-1}.
\end{equation}
If we write steps R and C together as a recursion on one of the
matrices~$\bA$ or~$\bB$, say, on the column space matrix~$\bA$,
we~get
\begin{equation}\label{eq:power}
\bA^{(k)}=\bW_r\bW_r^T{\bA^{(k-1)}}
\left({\bA^{(k-1)}}^T\bW_r\bW_r^T{\bA^{(k-1)}}\right)^{-1}{\bA^{(k-1)}}^T{\bA^{(k-1)}},
\end{equation}
which shows that, in the no missing data case, our RC~algorithm is
in fact implementing the application of the {\it power
method}~\cite{golub96} to the matrix~$\bW_r\bW_r^T$ (the factor
$(\bA^T\bW_r\bW_r^T\bA)^{-1}\bA^T\bA$ is the normalization). The
power method has been widely used to avoid the computation of the
entire SVD when fitting rank deficient matrices to complete
observations. We will see that, even when there is missing data,
both R- and C- steps admit closed-form solution and the overall
algorithm generalizes the  power method in a very simple way.

\subsubsection{R-step---estimation of the row space}

For known~$\bA$, the minimization of~(\ref{eq:minab}) wrt~$\bB$
can be rewritten in terms of each of the~$P_r$ columns~$\{\bb_p,
p=1,\ldots, P_r\}$ of~$\bB$,
\begin{equation}\label{eq:minbn}
\bb_p=\arg\min_{\widetilde{\bb}_p}
\left\|\left(\bw_p-\bA\widetilde{\bb}_p\right)\odot\bm_p\right\|_F
,
\end{equation}
where $\bw_p$ and~$\bm_p$ denote the $p^{\scriptsize\mbox{th}}$
columns of $\bW_r$ and~$\bM_r$, respectively. Exploiting the
structure of the binary column vector~$\bm_p$, we now re-arrange
the minimization in~(\ref{eq:minbn}) in such a way that its
solution becomes obvious. First, we note that~$\bm_p$
in~(\ref{eq:minbn}) is just selecting the entries of the error
vector~$(\bw_p-\bA\widetilde{\bb}_p)$ that affect the error norm.
Then, by making explicit that selection in terms of the entries
of~$\bw_p$ that contain known data and the corresponding relevant
entries of~$\bA$, we rewrite~(\ref{eq:minbn})~as
\begin{equation}\label{eq:minbn2}
\bb_p=\arg\min_{\widetilde{\bb}p}
\left\|\bw_p\odot\bm_p-\left(\bA\odot\bm_p\bone_{1\times
4}\right)\widetilde{\bb}_p\right\|_F .
\end{equation}
Note that $\bm_p\bone_{1\times 4}$ in~(\ref{eq:minbn2}) is simply
a $2F\!\times\!4$~matrix with all $4$~columns equal to~$\bm_p$.

The minimization in~(\ref{eq:minbn2}) is now clearly expressed as
a linear LS problem. Its solution~$\bb_n$ is then obtained by
using the pseudoinverse of matrix $(\bA\odot\bm_p\bone_{1\times
4})$,
\begin{equation}
\bb_p=\left[ \left(\bA\odot\bm_p\bone_{1\times
4}\right)^T\left(\bA\odot\bm_p\bone_{1\times 4}\right)\right]^{-1}
\left(\bA\odot\bm_p\bone_{1\times 4}\right)^T
\left(\bw_p\odot\bm_p\right) ,
\end{equation} which is simplified by omitting repeated binary
maskings,
\begin{equation} \label{eq:bgivena}
\bb_p=\left[\bA^T\left(\bA\odot\bm_p\bone_{1\!\times\!
4}\right)\right]^{-1}\bA^T \left(\bw_p\odot\bm_p\right).
\end{equation}
The set of~$P_r$ estimates $\{\bb_p, p=1,\ldots, P_r\}$ as
in~(\ref{eq:bgivena}) generalizes the well known pseudoinverse
LS~solution in (\ref{eq:pseudob}) to problems with missing data.

\subsubsection{C-step---estimation of the column space}

Given~$\bB$, the estimate of each row~$\ba_f$ of the column space
matrix~$\bA$ is obtained by proceeding in a similar way as in the
R-step. We get, for each $f=1,\ldots, 2F$,
\begin{equation} \label{eq:agivenb}
\ba_f=\left(\bw_f\odot\bm_f\right)\bB^T
\left[\left(\bB\odot\bone_{4\!\times\!
1}\bm_f\right)\bB^T\right]^{-1},
\end{equation}
where in this case, for commodity, lowercase boldface letters
denote rows rather than columns.

\section{Experiments---rank reduction with missing data}\label{sec:exp1}

Since approximating rank deficient matrices from incomplete
observations is a key step in recovering 3-D structure, we start
by describing experiments that illustrate the behavior of the EM
and RC algorithms just described and demonstrate their good
performance.

\subsection{2$\times$2 matrices}

We start by a simple case that allows an illustrative graphical
representation---approximating $2\times 2$ rank~1 matrices from
observations $\bW_r$ that miss one of the entries. In this case,
the error measured by the Frobenius norm in
expressions~(\ref{eq:ll}) and~(\ref{eq:minab}), can be expressed
in terms of a single parameter~$\theta$. In fact, let the $2\times
2$ rank~1 matrix~$\widetilde{\bW}$ be written in terms of its
column and row spaces as in~(\ref{eq:ab}),
\begin{equation}\label{eq:ab2x2r1}
\widetilde{\bW}=\ba_{2\times 1}\bb_{1\times 2}\in\cS_1.
\end{equation}
Without loss of generality, impose that the row vector~$\bb$ has
unit norm and write it in terms of a row angle~$\theta$,
\begin{equation}
\bb=\left[\begin{array}{cc}\cos\theta &
\sin\theta\end{array}\right].\label{eq:theta}
\end{equation}
Now denote the minimum of (\ref{eq:minab}) wrt the column
space~$\ba$ for fixed row space~$\bb$, {\it i.e.}, for
fixed~$\theta$, by ${\ba}(\bW_r,\theta)$, given
by~(\ref{eq:agivenb}). The error in~(\ref{eq:ll}) and
(\ref{eq:minab}) is then rewritten as a function of~$\theta$~as
\begin{equation}\label{eq:errtheta}
\mbox{error}(\theta)=\left\|\left(\bW_r-{\ba}(\bW_r,\theta)
\left[\begin{array}{cc}\cos\theta &
\sin\theta\end{array}\right]\right)\odot\bM_r\right\|_F.
\end{equation}
Note that for any set of three entries of a $2\times 2$ matrix,
there is always a value for the forth entry that makes the rank of
the matrix equal to one, {\it i.e.}, there is always a $2\times 2$
rank~1 matrix~$\widetilde{\bW}$ that fits exactly the observed
entries of~$\bW_r$. Thus, we have $\min\mbox{error}(\theta)=0$.

The examples in the sequel illustrate the impact of the
initialization on the behavior of the EM and RC algorithms with
experiments that use the following ground truth rank~1
matrix~$\widetilde{\bW}$, observation mask~$\bM_r$, and
observation~$\bW_r$:
\begin{equation}\label{eq:2x2_W_initial}
  \widetilde{\bW}=\left[\begin{array}{cc}
    -1&-1.95\\
    2&3.9
\end{array}\right]\in\cS_1,\qquad
\bM_r=\left[\begin{array}{cc}
    1&1\\
    1&0
\end{array}\right],\qquad
  \bW_r=\left[\begin{array}{cc}
    -1&-1.95\\
    2&?
\end{array}\right],
\end{equation}
where ``$?$" represents the unobserved entry~$w_{22}$ of the
observation matrix~$\bW_r$.

\subsection{Typical good behavior of EM and RC}

\begin{figure} [htb]
 \centerline{\psfig{figure=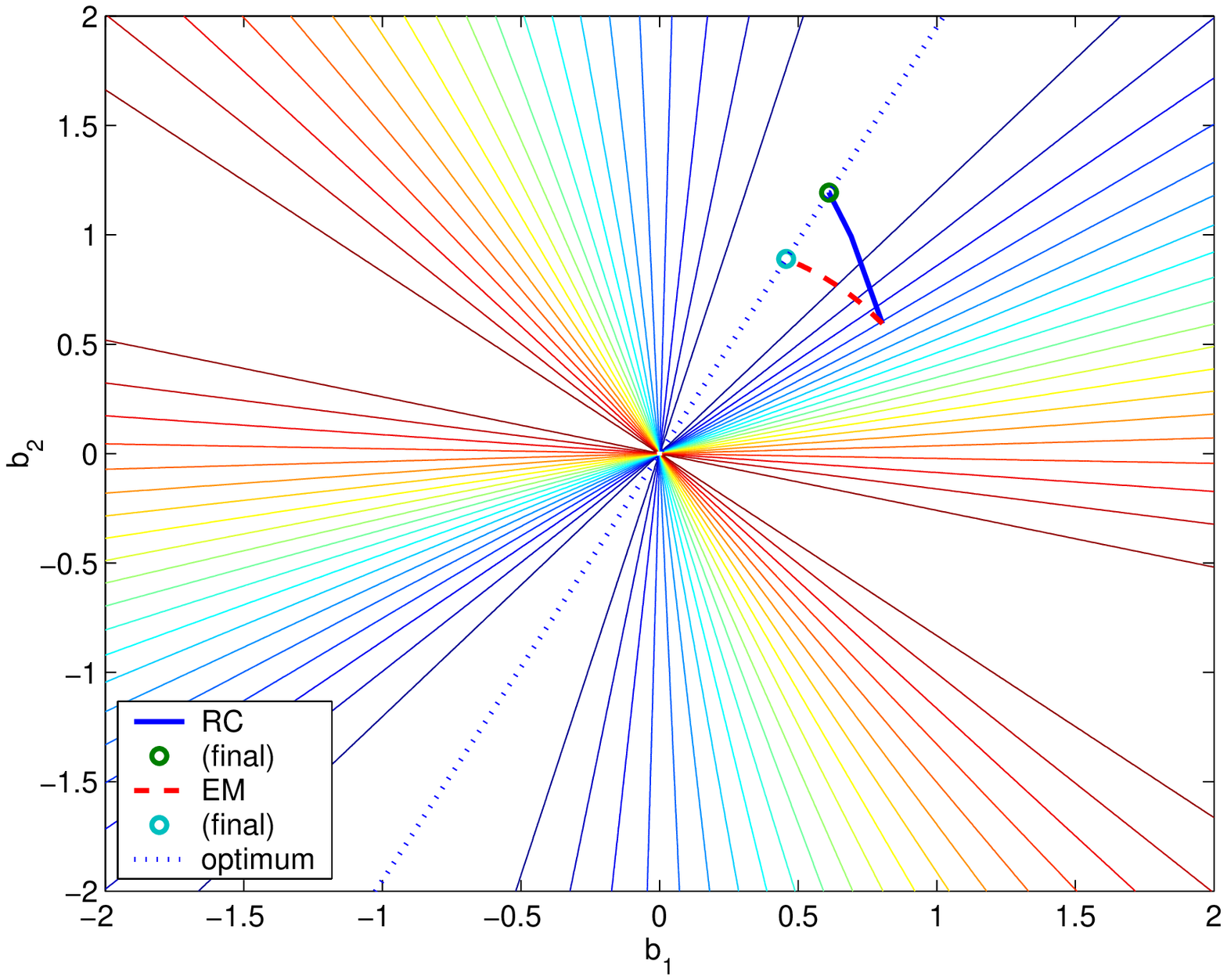,width=7cm}\hspace*{1cm}
 \psfig{figure=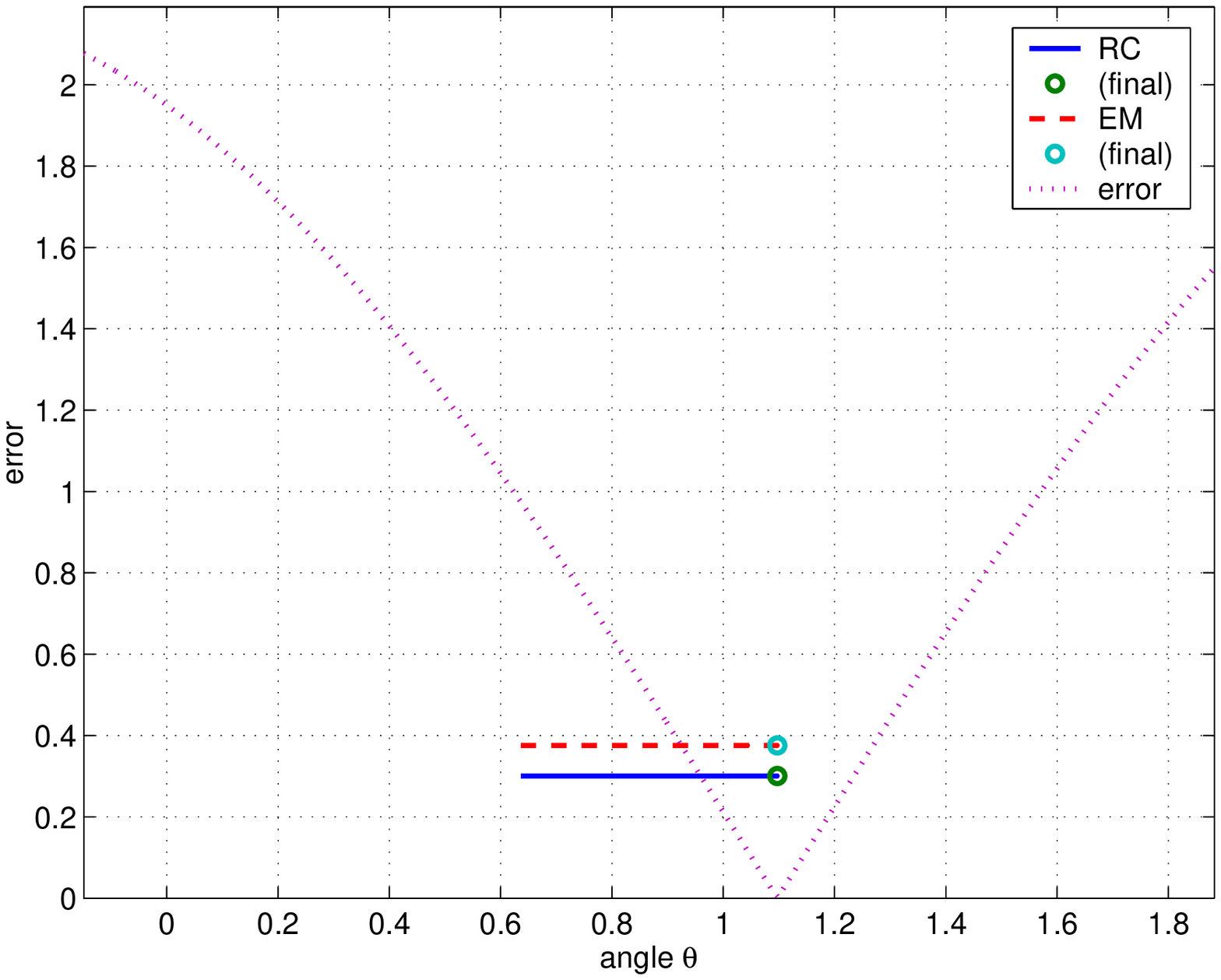,width=7cm}}
  \caption{Typical good behavior. Left: trajectories of the estimates of the iterative algorithms
  EM and RC. Right: error function.
  \label{fig:i0_1}}
\end{figure}

Using the initial guess
\begin{equation}\label{eq:winit0}
  \widehat{\bW_r}^{(0)}=\left[\begin{array}{cc}
    -1&-1.95\\
    2&0
\end{array}\right],
\end{equation}
we describe the evolution of the estimates~$\widehat{\bW_r}^{(k)}$
of the rank~1 matrix by plotting two equivalent representations
of~$\widehat{\bW_r}^{(k)}$: {\bf i)} its row
space~$\bb=[b_1,b_2]$; and {\bf ii)} the corresponding angle
$\theta$ as defined in~(\ref{eq:theta}). The left plot of
Fig.~\ref{fig:i0_1} shows the level curves of the error function
as function of the row vector $\bb=[b_1,b_2]$, superimposed with
the evolution of the estimates of~$\bb$ for EM (dashed line) and
RC (solid line). In this plot, the dotted line (optimum) are the
row vectors that lead to zero estimation error. The right plot of
Fig.~\ref{fig:i0_1} represents the same error function, now as a
function of~$\theta$, as defined in~(\ref{eq:errtheta}), (dotted
line) superimposed with the locations of the~$\theta$ estimates
for EM (dashed line) and RC (solid line).

From the left plot of Fig.~\ref{fig:i0_1}, we see that both EM and
RC trajectories start at the same point (due to the equal
initialization) and converge to points in the optimal line. As
expected, the EM estimates of the row space vector have constant
unit norm (due to the normalization in the SVD) while the RC
estimates don't. The good behavior of the algorithms is confirmed
by the right plot of Fig.~\ref{fig:i0_1} that shows that both
algorithms converge to a value of~$\theta$ that makes
$\mbox{error}(\theta)=0$, {\em i.e.}, that minimizes
$\mbox{error}(\theta)$.

\begin{figure} [htb]
 \centerline{\psfig{figure=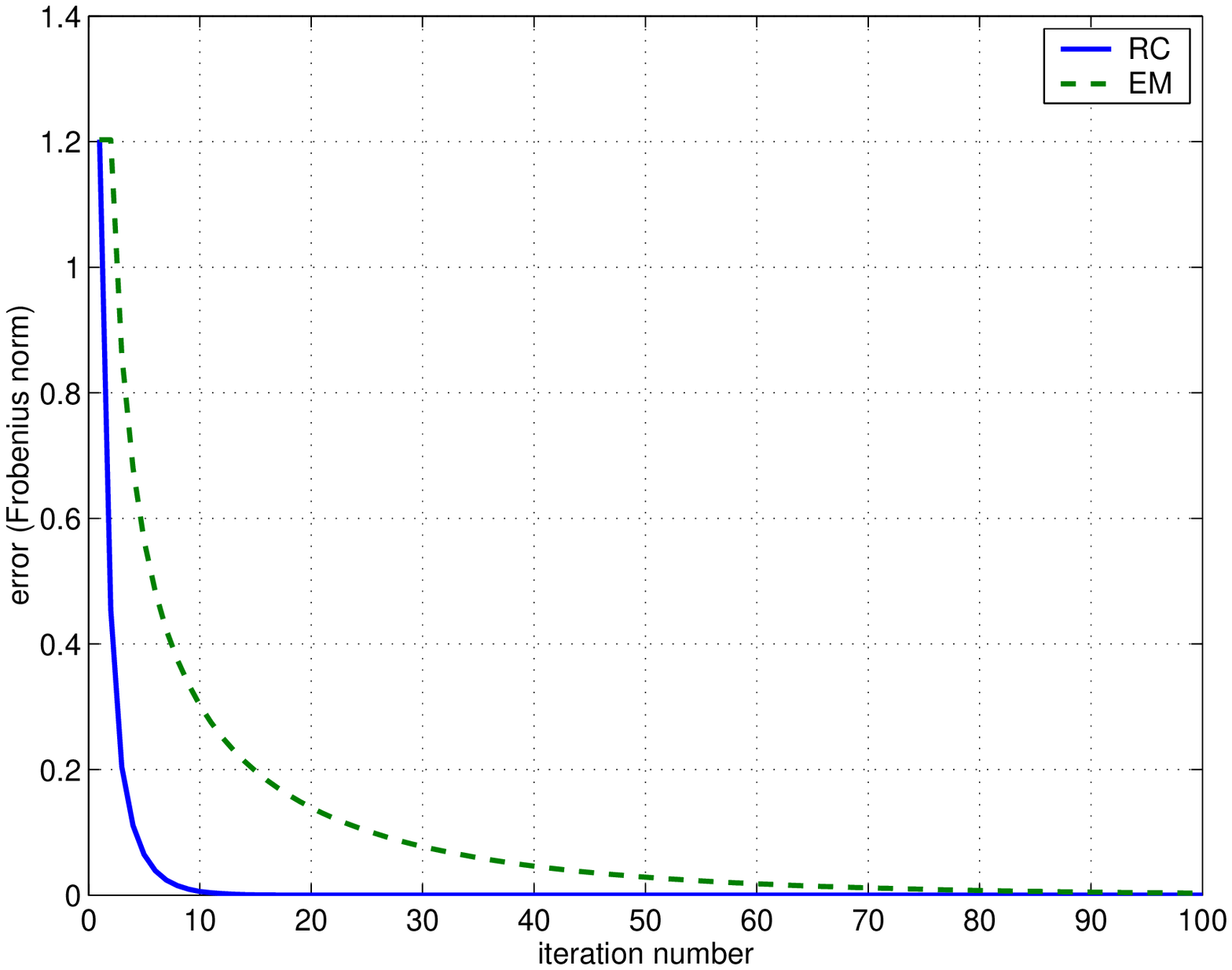,width=7cm}\hspace*{1cm}
 \psfig{figure=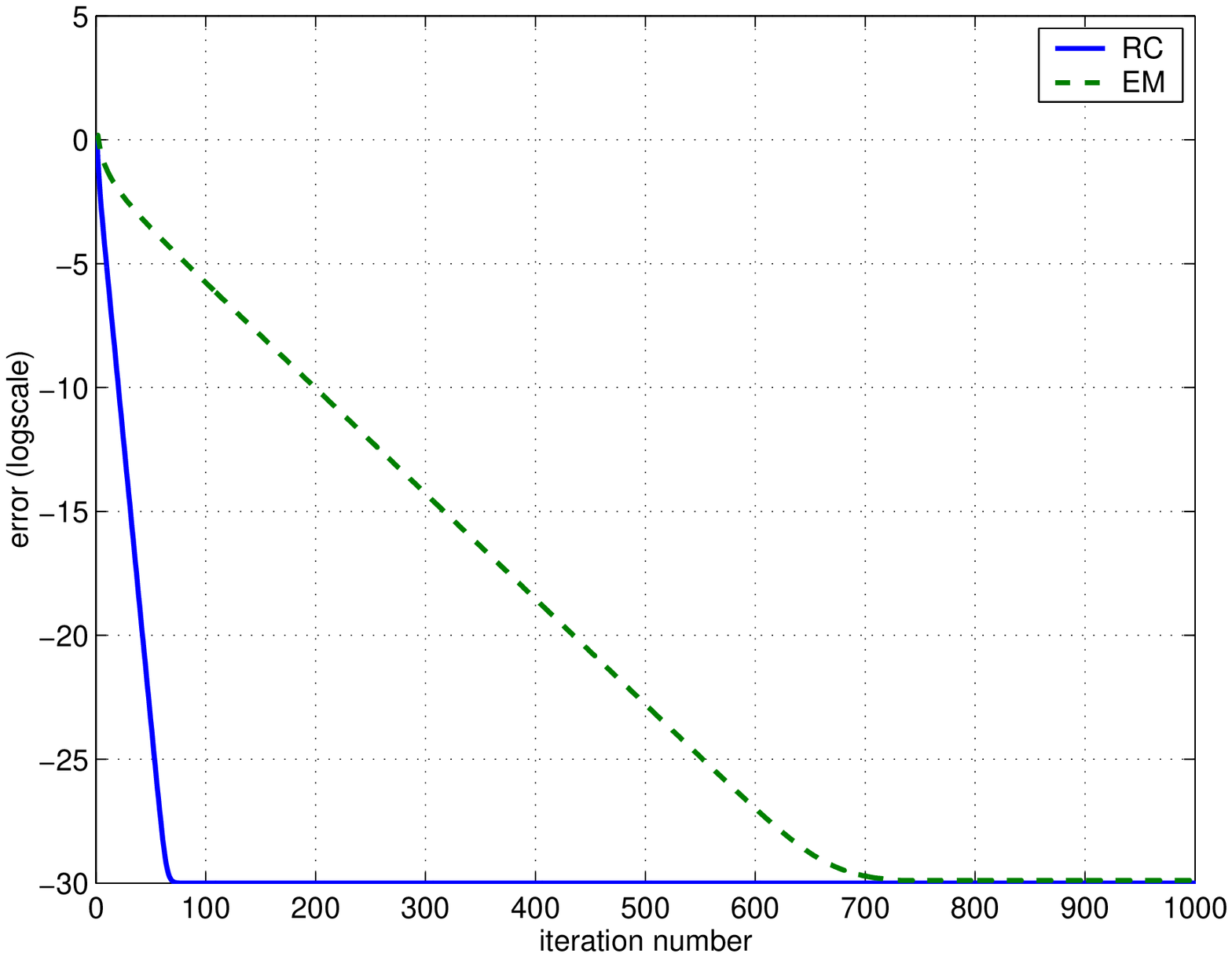,width=7cm}}
  \caption{Typical good behavior. Left: estimation error. Right: the same in logscale.
  \label{fig:i0_2}}
\end{figure}

To evaluate the convergence speed, we show in Fig.~\ref{fig:i0_2}
the evolution of the estimation error along the iterative process
for both algorithms (in the left plot, linear scale, in the right
one, logarithmic scale). From Fig.~\ref{fig:i0_2}, we see that RC
converges in a smaller number of iterations than EM. Our
experience with larger matrices in practical applications have
confirmed the faster convergence of RC.

\subsection{Large entries in rows or columns that miss data}

\begin{figure} [htb]
 \centerline{\psfig{figure=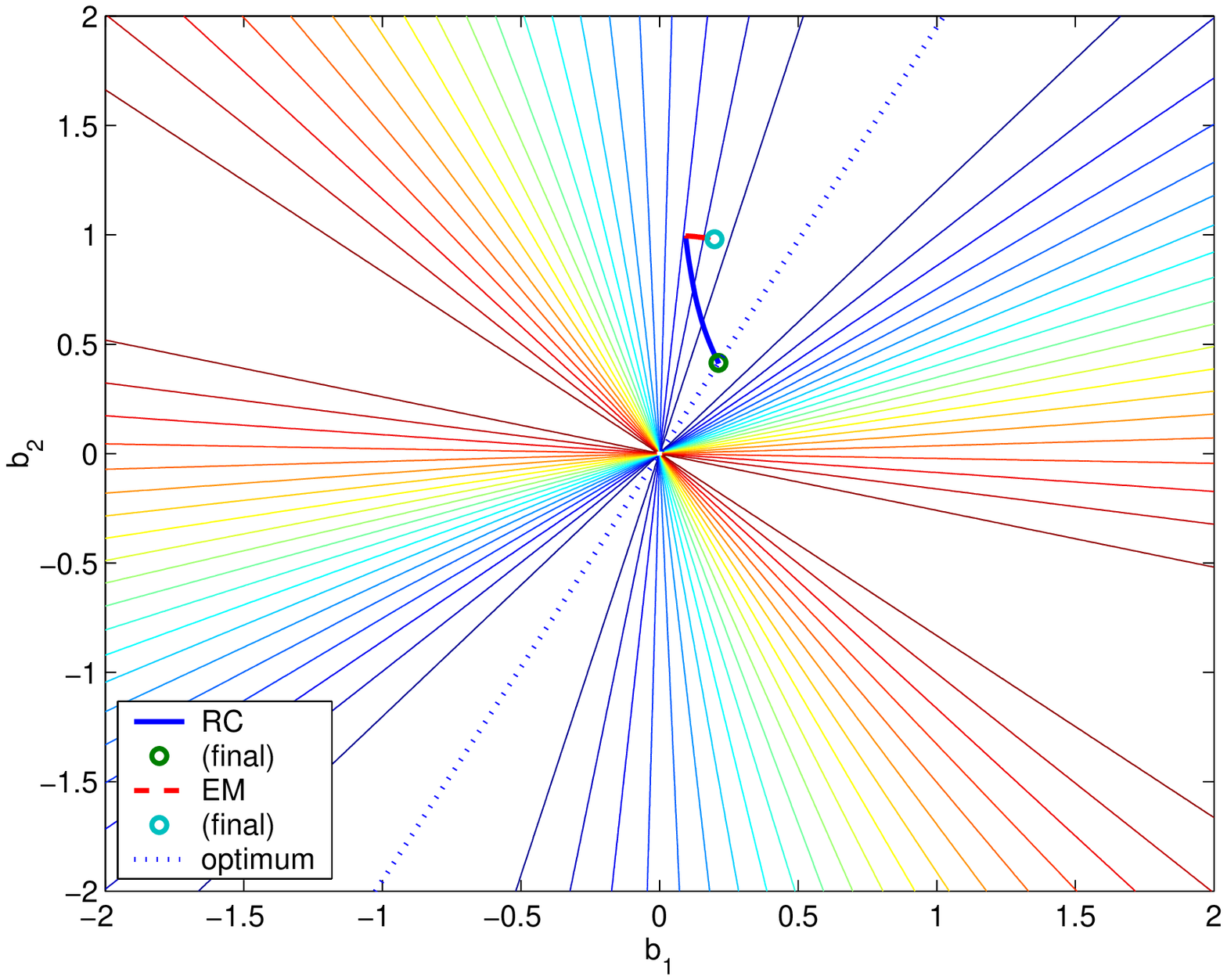,width=7cm}\hspace*{1cm}
 \psfig{figure=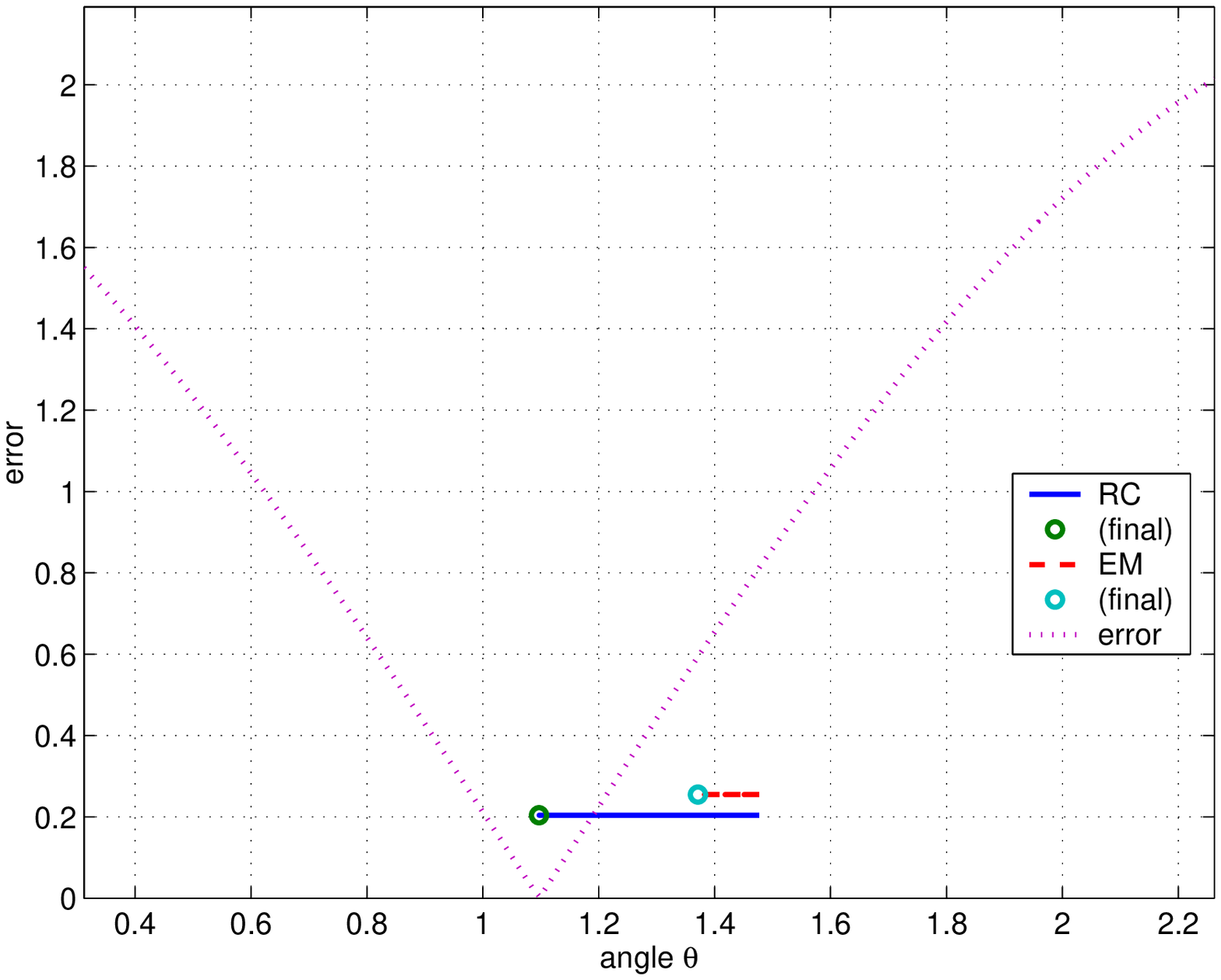,width=7cm}}
  \caption{Large initialization. Left: EM and RC trajectories.
  Right:error function.
  \label{fig:i22_1}}
\end{figure}

We now illustrate a drawback of the EM algorithm. Using the same
data and the initial estimate
\begin{equation}\label{eq:winitlarge}
  \widehat{\bW_r}^{(0)}=\left[\begin{array}{cc}
    -1&-1.95\\
    2&22
\end{array}\right],
\end{equation}
we get the first $1000$ iterations of Fig.~\ref{fig:i22_1}, which
is as described above for Fig.~\ref{fig:i0_1}.

\begin{figure} [htb]
 \centerline{\psfig{figure=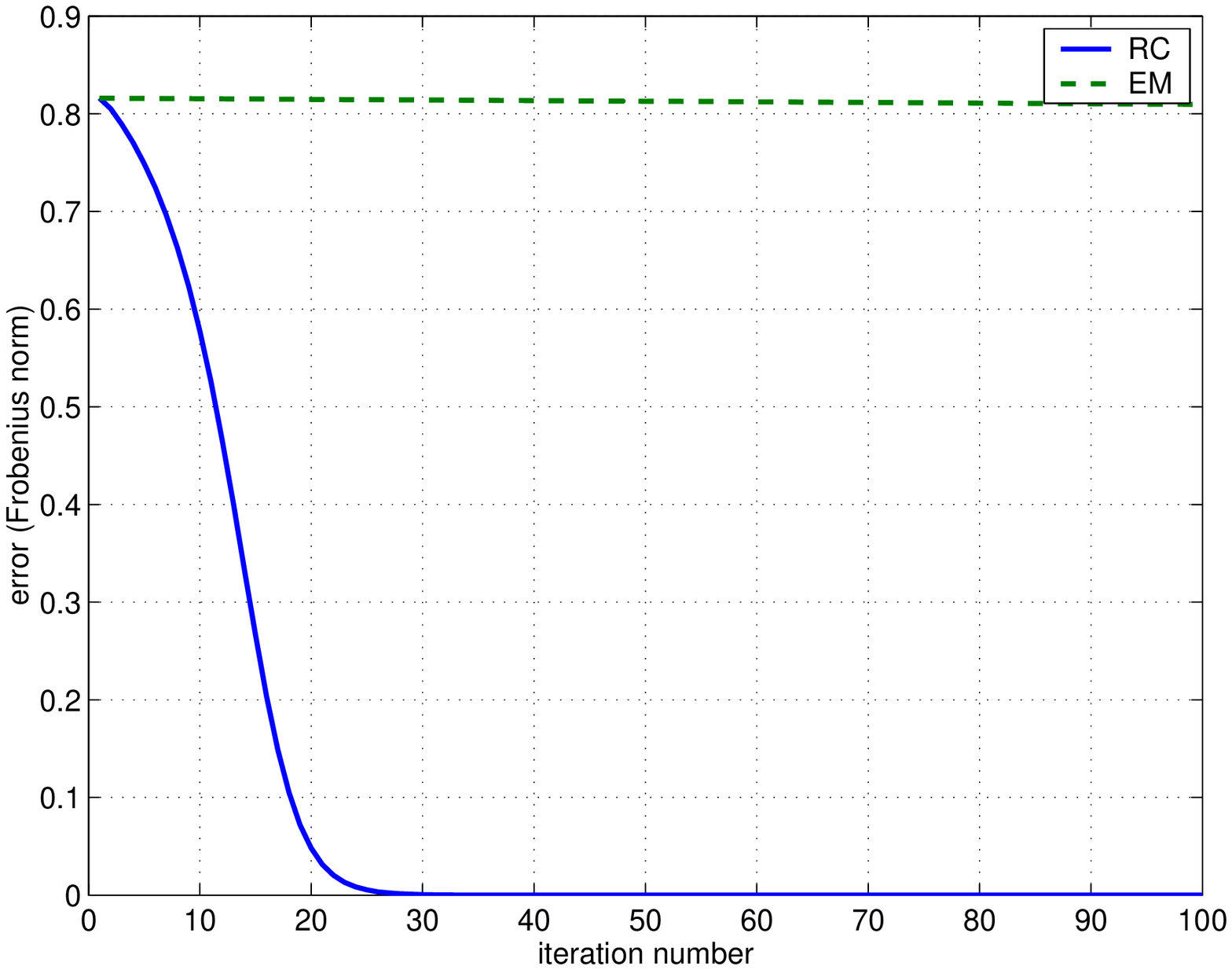,width=7cm}\hspace*{1cm}
 \psfig{figure=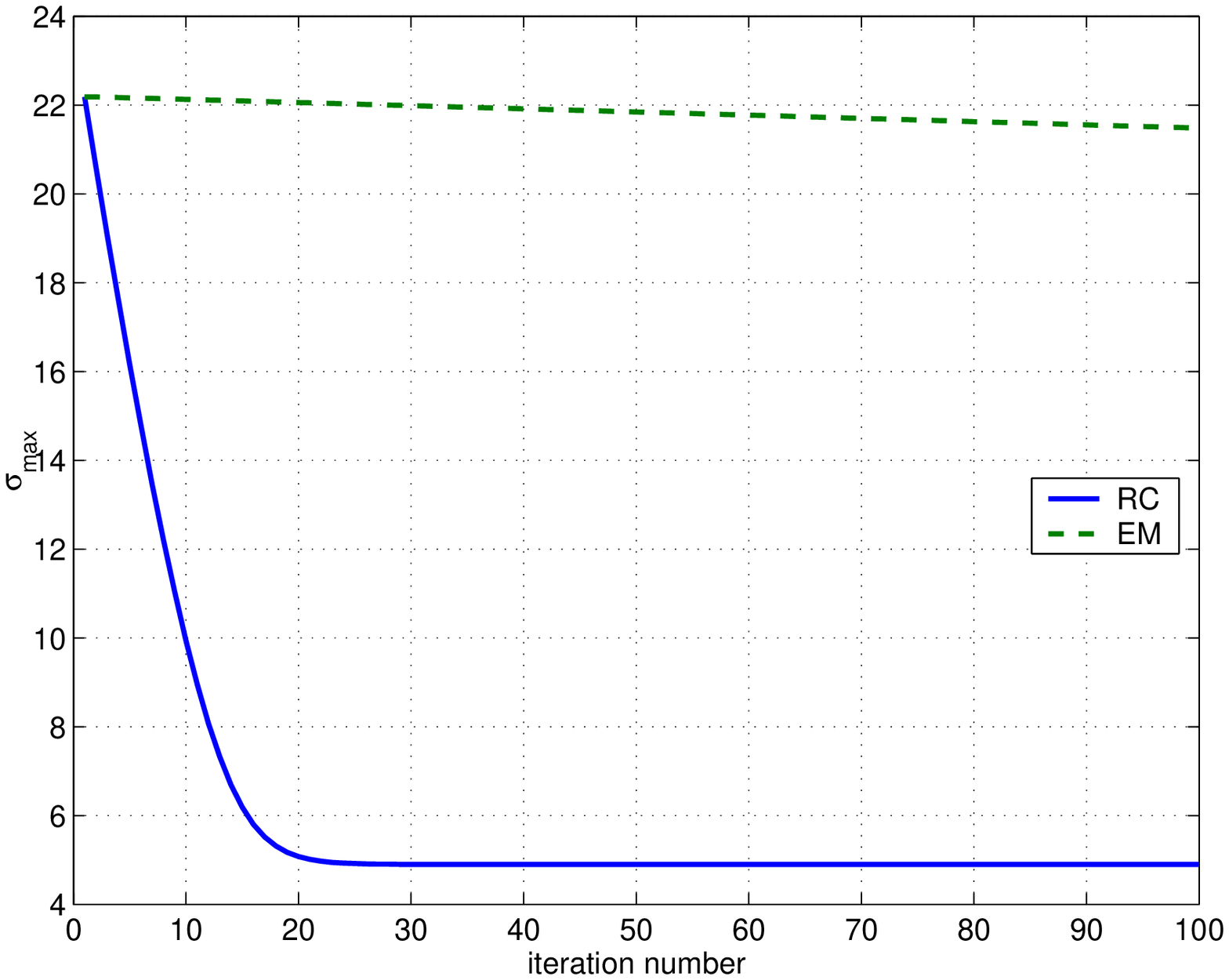,width=7cm}}
  \caption{Large initialization. Left: estimation error. Right:
  largest singular value of~$\widehat{\bW}_k$, the estimate at the $k-$th iteration.
  \label{fig:i22_2}}
\end{figure}

From the left plot of Fig.~\ref{fig:i22_1}, we see that, while RC
converges to the optimal line, the estimates given by the EM
almost don't change along the iterative process. This can also be
seen in the right plot of Fig.~\ref{fig:i22_1}, which shows that
RC converges to the $\theta$ such that $\mbox{error}(\theta)=0$,
while EM, after $1000$ iterations is still far from
$\arg\min\mbox{error}(\theta)$. The left plot of
Fig.~\ref{fig:i22_2} shows the evolution of the estimation errors.
See that while the error of RC converges to zero in a few
iterations, the error of EM almost doesn't decrease during the
first $100$ iterations.

The bad behavior of EM is due to the large initial guess
for~$w_{22}$ (note that $\widehat{w}^{(0)}_{22}$ is large when
compared with the known entries of~$\bW_r$). Remember that EM
starts by estimating the rank~1 matrix that best matches the
initial guess $\widehat{\bW_r}^{(0)}$. This rank~1 matrix is the
matrix associated with the largest singular value of
$\widehat{\bW_r}^{(0)}$, which is highly constrained by the large
spurious entry $\widehat{w}^{(0)}_{22}=22$ (note that while the
singular values of the ground truth matrix~$\widetilde{\bW}$ are
$\sigma_1(\widetilde{\bW})\!\simeq\!4.9$ and \linebreak
$\sigma_2(\widetilde{\bW})\!=\!0$, the singular values of the
initial guess $\widehat{\bW_r}^{(0)}$
are~$\sigma_1(\widehat{\bW_r}^{(0)})\!\simeq\! 22.2$ and
\linebreak $\sigma_2(\widehat{\bW_r}^{(0)})\!\simeq\!0.8$ due to
the large entry~$\widehat{w}^{(0)}_{22}=22$). Then, EM replaces
the known entries of $\bW_r$ in the new estimate (obtaining thus
an estimate that is very close to the initial guess) and repeats
the process. To better illustrate this very slow convergence of
EM, we represent, in the right plot of Fig.~\ref{fig:i22_2}, the
evolution of the largest singular value of the estimate
$\widehat{\bW_r}^{(k)}$ for both algorithms. We see that while, as
expected, the largest singular value of the RC estimates converges
to the largest singular value of the solution~$\widetilde{\bW}$,
$\sigma_1(\widetilde{\bW})\!\simeq\!4.9$, the largest singular
value of the first $100$ iterations of the EM estimates changes
very slowly from its initial
value~$\sigma_1(\widehat{\bW_r}^{(0)})\!\simeq\! 22.2$.

The behavior just described also happens in situations other than
the initial guesses of the unknown entries being too large when
compared to the other entries. In fact, we observed the same
behavior in situations where the observation matrix had large
entries in rows or columns that contained missing entries. In
these situations, due to those large entries, even small values
for initial guess of the unknown entries had large impact on the
row and column singular vectors associated with the large singular
values that determined the best rank deficient approximations to
the complete data matrices involved in EM. This lead to the same
kind of very slow convergence illustrated in Fig.~\ref{fig:i22_1}
and Fig.~\ref{fig:i22_2}.

\subsection{Large matrices}

We now present results of approximating matrices likely to appear
when processing real video sequences. We tested the EM and RC
algorithms with noisy partial observations of rank~4 matrices. We
used matrices with dimensions ranging from $2\times 2$
to~$200\times 200$. The percentage of missing entries ranged from
$10\%$ to $80\%$. We studied the behavior of the EM and RC
algorithms in terms of the observation noise power and the amount
of missing data, and the impact of the initialization.

\begin{figure} [htb]
 \centerline{\psfig{figure=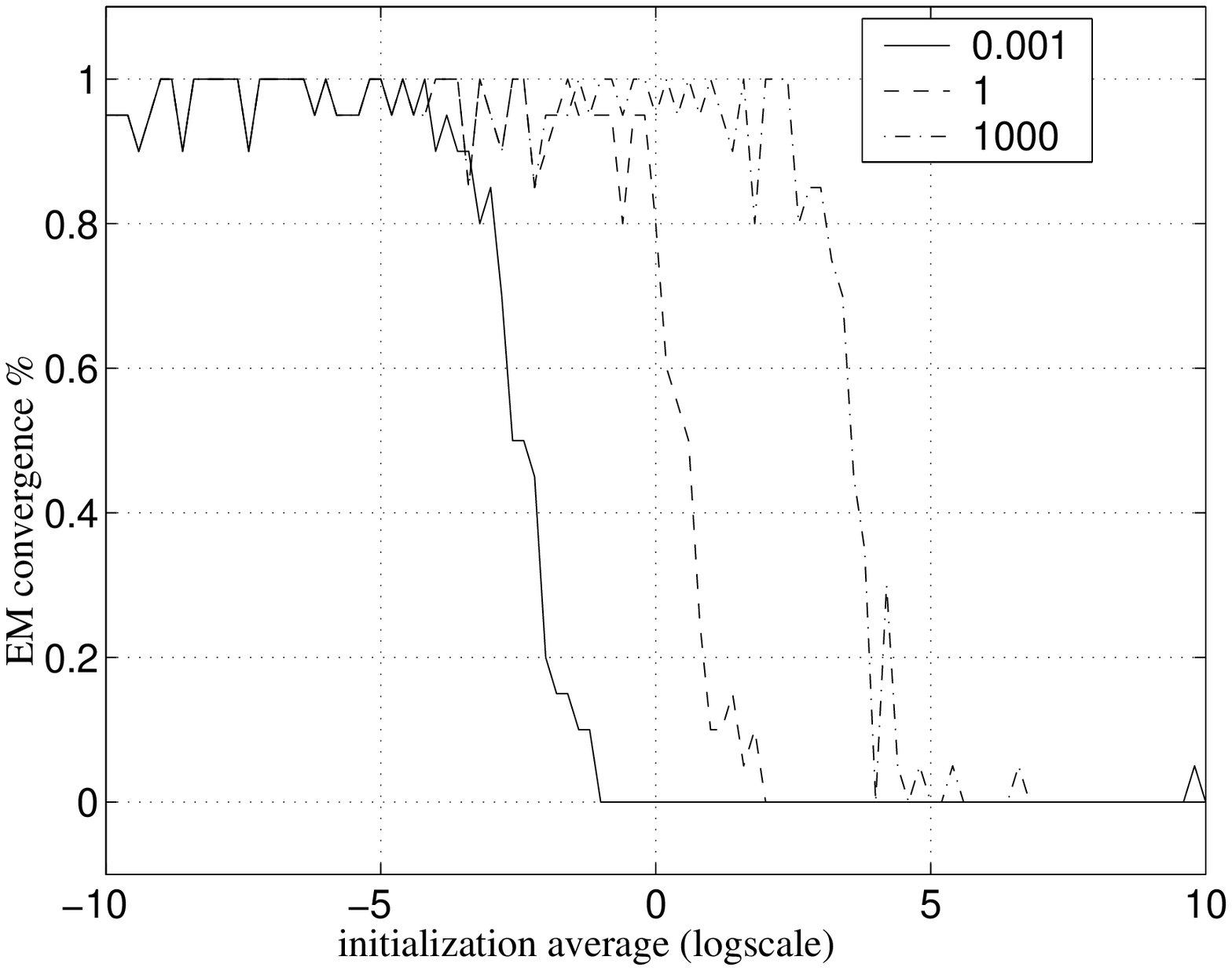,width=7cm}
 \hspace*{1cm}
 \psfig{figure=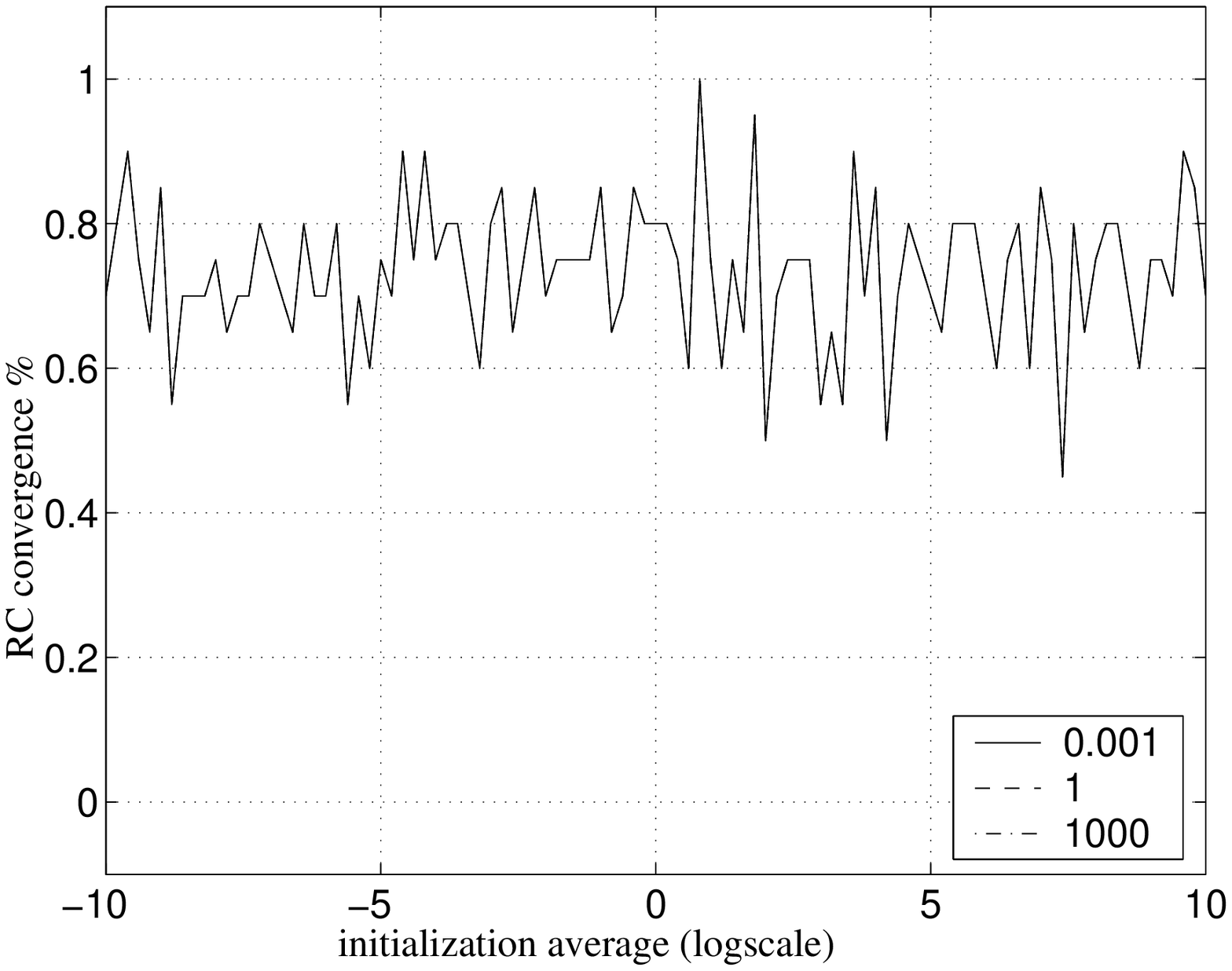,width=7cm}}
 \caption{Percentage of convergent experiments as a function of the mean value of the initial
random guesses for the unknown entries (ground truth matrices
whose entries have mean~$10^{-3}$, $1$, and~$10^3$). Left:  EM
algorithm. Right:  RC algorithm.
  \label{fig:p_conv}}
\end{figure}

We initialized both EM and RC with random values for the unknown
entries. To illustrate the influence of the initialization on the
convergence of the algorithms, Fig.~\ref{fig:p_conv} shows the
percentage of experiments that converged (to an estimate close
enough to the ground truth) in less than 100 iterations, as a
function of the mean value of the random guesses for the unknown
entries. Although representative for the entire range of
experiments done, the results in the plots of
Fig.~\ref{fig:p_conv} were obtained with noisy observations of
$24\times 24$ rank~4 matrices that missed $70\%$ of their entries.
The left plot of Fig.~\ref{fig:p_conv} shows three lines, each
obtained by using EM with data generated with a ground truth
matrix whose elements had mean~$0.001$, $1$, and~$1000$. The
percentages of convergence for the RC algorithm are in the right
plot.

The left plot of Fig.~\ref{fig:p_conv} shows that the EM algorithm
converges almost always if the mean value of the initial guesses
for the missing entries is smaller than the mean value of the
observed entries. When we increase the values of the initial
estimates of the missing entries, the percentage of convergence
decreases abruptly, becoming close to zero when those values
become much larger than the ones of the observed entries. This is
in agreement with the behavior illustrated in the example of
Fig.~\ref{fig:i22_1}. In opposition, the right plot of
Fig.~\ref{fig:p_conv} shows that the behavior of the RC algorithm
is somewhat independent of the order of magnitude of the
initialization. The experiments that lead to a non-convergent
behavior of RC were such that the matrices whose inverse is
computed in~(\ref{eq:agivenb}) and~(\ref{eq:bgivena}) were close
to singular. We thus conclude that it is very important in
practical applications to provide good initial estimates for both
EM and RC algorithms.

Finally, we note that the relevance of a good initialization goes
behind avoiding non-convergent behavior. In fact, we observed that
both the amount of missing data and the noise level have strong
impact on the algorithm's convergence speed. Thus, when dealing
with large percentages of missing entries and high levels of
noise, as it may arise in practice, a better initialization not
only improves the chance of a convergent behavior but also leads
to a faster convergence.

\subsection{Heuristic initialization}

We now use an initial guess~$\widehat{\bW_r}^{(0)}$ obtained by
combining the column and row spaces given by the SVDs of the known
submatrices of~$\bW_r$ as described in the appendix. In our tests,
with this simple initialization procedure, $100\%$ of the runs of
EM and RC converged to the ground truth matrix (mean square
estimation error smaller than the observation noise variance) in a
very small number of iterations, typically less than 10, even for
high levels of noise.

\begin{figure} [htb]
  \centerline{\psfig{figure=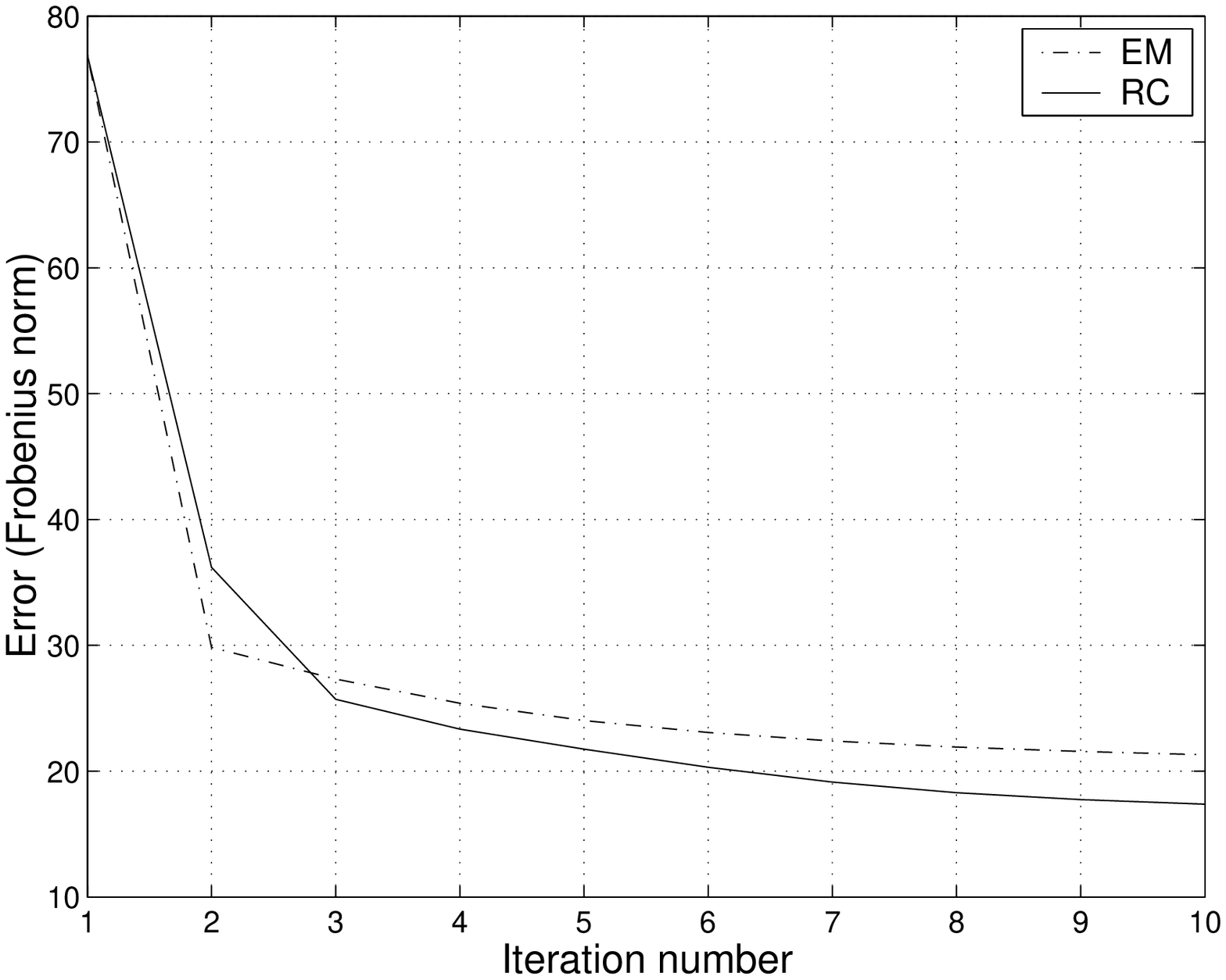,height=6cm}\hspace*{2cm}
  \psfig{figure=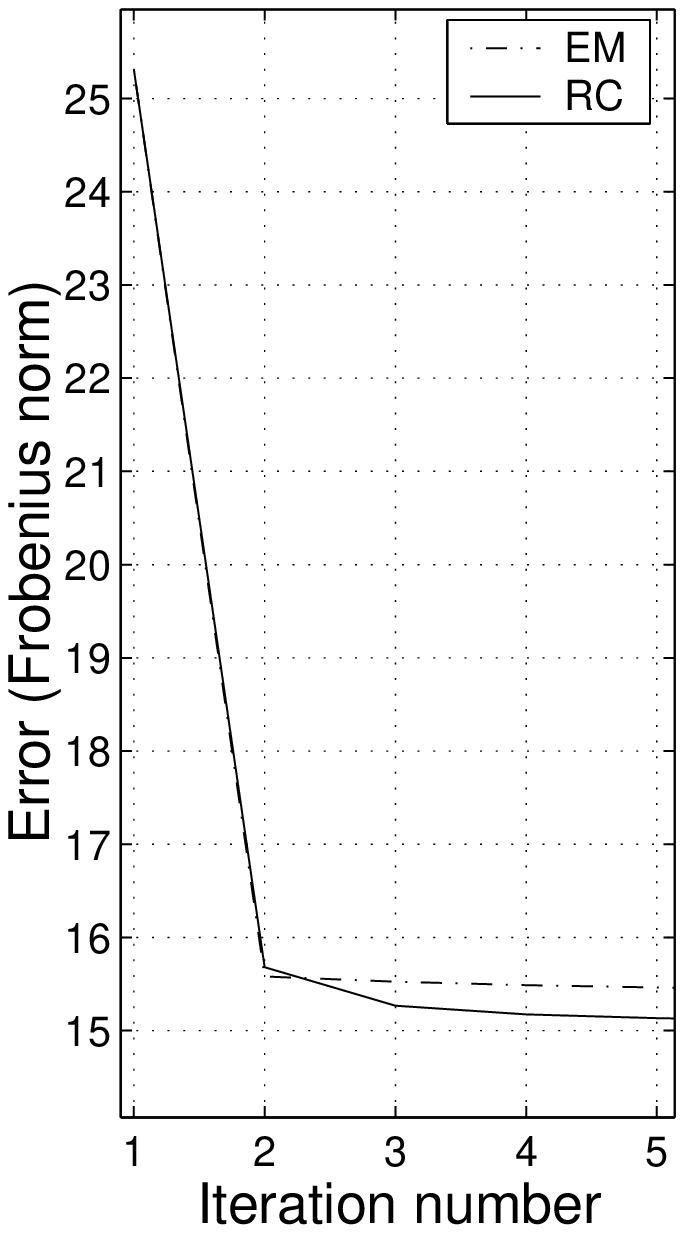,height=6cm}}
  \caption{Evolution of the estimation error for a $40\times 40$
  rank deficient matrix with $30\times 30$ missing entries,
   with a random initialization (left plot) and with the heuristic
  initialization described in the appendix (right plot).\label{fig:tudo}}
\end{figure}

To illustrate the impact of a better initialization, we use a
noisy observation~$\bW_r$ of a $40\times 40$~rank deficient
matrix~$\widetilde{\bW}$ with $900$ missing entries (observation
noise variance~$\sigma^2=1$). Figure~\ref{fig:tudo} shows plots of
the evolution of the estimation error, measured by the Frobenius
norm, for both the EM and RC algorithms. While for the left hand
side plot we used a random initialization, for the one in the
right we used the complete procedure, {\it i.e.}, we initialized
the process by using the method described in the appendix. Since
the error for the ground truth matrix~$\widetilde{\bW}$ is given
by
$\left\|(\widetilde{\bW}-\bW_r)\odot\bM_r\right\|_F\simeq\sigma\sqrt{40^2-900}\simeq
26.5$, we see from the plots of figure~\ref{fig:tudo} that both
the EM and RC algorithms provide good estimates. From the right
hand side plot, we conclude that the initial guess provided by the
heuristic method in the appendix enables a much faster convergence
(in 2 or 3~iterations) to a solution with lower error.


\subsection{Sensitivity to the noise}

\begin{figure} [htb]
 \centerline{
 \psfig{figure=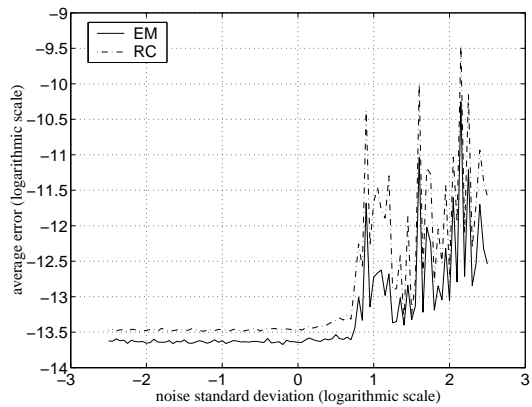,width=7cm}
 }
 \caption{Estimation errors for the algorithms EM and RC as functions of the noise standard
 deviation.\label{fig:err_w_noise}}
\end{figure}

The plot of Fig.~\ref{fig:err_w_noise} represents the average
estimation error per entry
after 20 iterations of EM and RC algorithms, for noisy
observations of a $24\times 24$ rank 4 matrix~$\widetilde{\bW}$,
with~$70\%$ missing data, as a function of the observation noise
standard deviation. We see that the average entry estimation
errors after $20$~iterations are below~$10^{-8}$ for noise
standard deviation ranging from~$10^{-2.5}$ to~$10^{2.5}$ (the
mean value of the entries of the ground truth
matrix~$\widetilde{\bW}$ is~$1$). This shows that, with a simple
initialization procedure, both EM and RC algorithms converge to
the optimal solution, even for very noisy observations.
Furthermore, we conclude that the main impact of the observation
noise is on the EM and RC convergence speeds---the slightly higher
average error values on the right region of the plot of
Fig.~\ref{fig:err_w_noise} indicates that the estimates were still
converging to the optimal solution after $20$~iterations.

\subsection{Sensitivity to the missing data}
A relevant issue is the robustness of the rank reduction
algorithms to the missing data. Our experience showed that the
structure of the binary mask matrix~$\bM_r$ representing the known
data is more important than the overall percentage of missing
entries of~$\bW_r$. When recovering 3D~structure from video,
feature points enter and leave the scene in a continuous way, thus
the typical structure of~$\bM_r$ is as represented in the left
plot of Fig.~\ref{fig:matrix}, or, when considering re-appearing
features, as represented in the right plot of the same Figure. We
tested the EM and RC algorithms with several mask matrices~$\bM_r$
with this typical structure, with the heuristic initial guess. In
 these experiments, the algorithms converged always to the
optimal solution in a very small number of iterations,
independently of the percentage of missing data.

Similarly to the impact of the noise discussed above, the
percentage of missing entries has impact on the convergence speed.
In fact, we concluded that the larger is the portion of the matrix
that is observed, {\it i.e.}, the smaller is the percentage of
missing data, the lower is the estimation error after a fixed
number of iterations, {\it i.e.}, the faster is the convergence of
the iterative algorithms.




%

\subsection{Computational cost}

As referred above, the EM and RC algorithms converge in a very
small number of iterations when initialized by the heuristic
procedure described in the appendix. We now report an experimental
evaluation of the computational costs of each iteration of EM and
RC as functions of the observation matrix dimension.

\begin{figure} [htb]
 \centerline{\psfig{figure=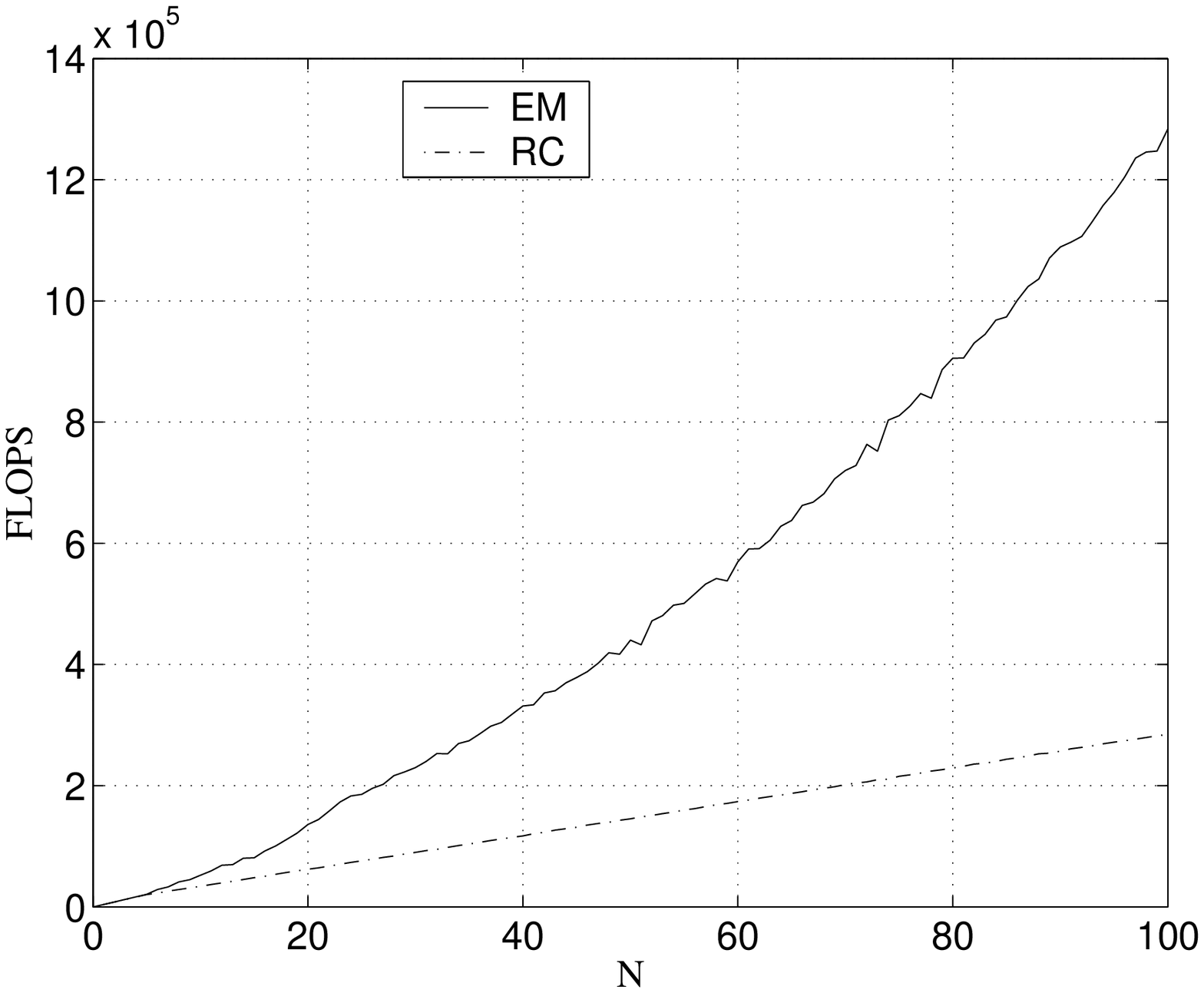,width=7cm}\hspace*{1cm}
 \psfig{figure=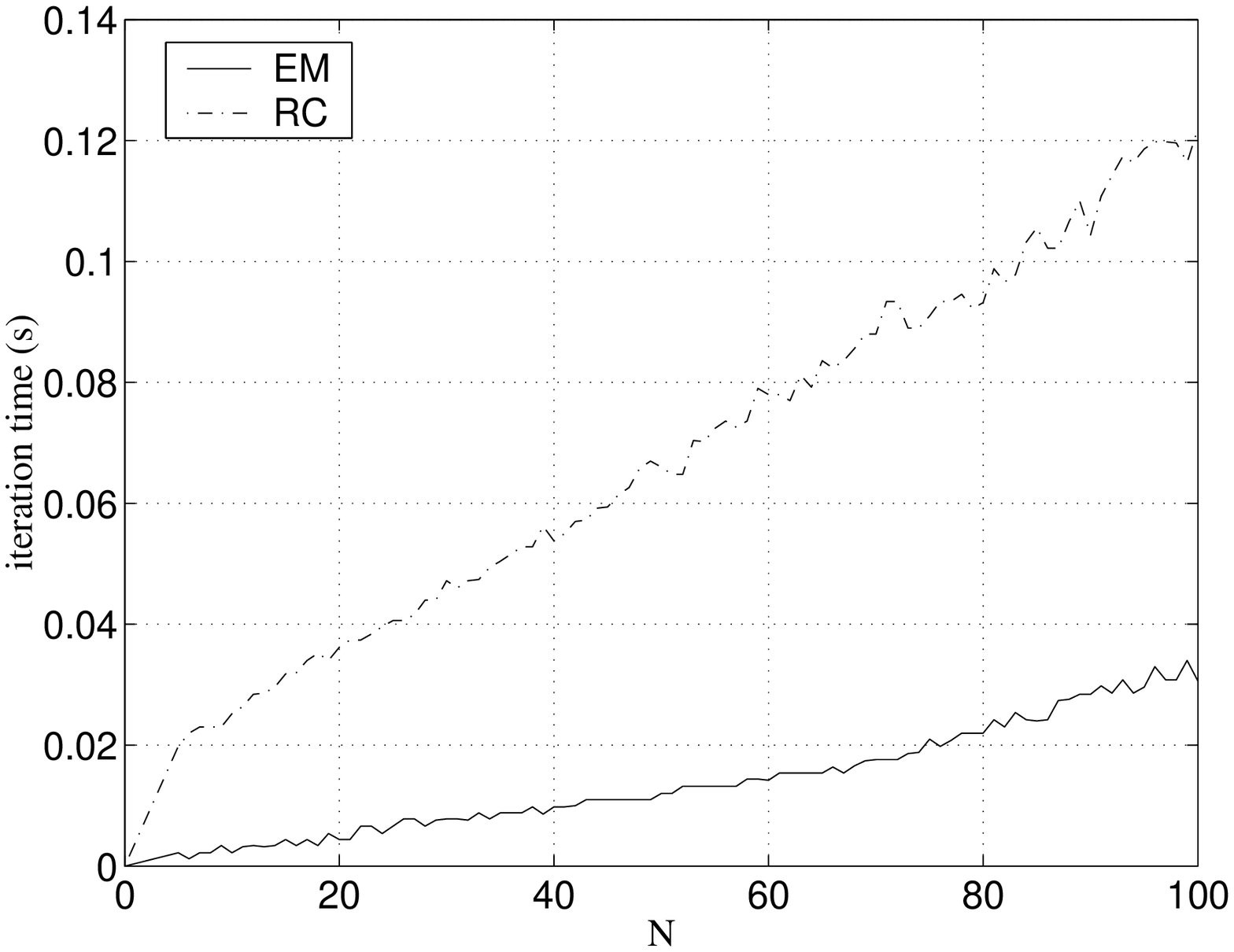,width=7cm}}
  \caption{Computational cost of each iteration of the algorithms EM and RC.
   Left: number of MatLab$^{\scriptsize\copyright}$ floating point operations~(FLOPS).
   Right: computation time.}\label{fig:comp}
\end{figure}

We used $N\times 24$ observation matrices with missing data
corresponding to a $(N-4)\times 20$ submatrix. The plots in
Fig.~\ref{fig:comp} represent the number of
MatLab$^{\scriptsize\copyright}$ floating point operations~(FLOPS)
and the computation time per iteration, as functions of~$N$. From
the left plot, we see that the number of FLOPS per iteration of
the EM algorithm is larger than one of the RC algorithm.
Furthermore, the FLOPS count for EM increases exponentially
with~$N$ (due to the SVD computation) while for RC it increases
linearly with~$N$.
Thus, although the computation times in the right plot of
Fig.~\ref{fig:comp} are smaller for EM than for RC (the reason
being the very efficient MatLab$^{\scriptsize\copyright}$
implementation of the SVD), we conclude that RC is computationally
simpler than EM. RC is even as simple as the methods to deal with
complete matrices, since the most efficient way to compute the SVD
is to use the power method~\cite{golub96} of which RC is a simple
generalization.


\section{Experiments---3-D models from video}\label{sec:exp2}

We now describe experiments that illustrate our method to recover
3-D rigid shape and 3-D motion from video.
%
%
Our results show: {\bf i)} how the iterative algorithms of
section~\ref{sec:missing} cope with video sequences with
occlusion; and {\bf ii)} how the global method we propose recovers
complete 3-D models. The experiments use both synthesized and real
video sequences. We also describe an application of the recovered
3-D models to video compression.

\subsection{Self-occlusion---ping-pong ball video sequence}

\begin{figure} [htb]
 \centerline{\psfig{figure=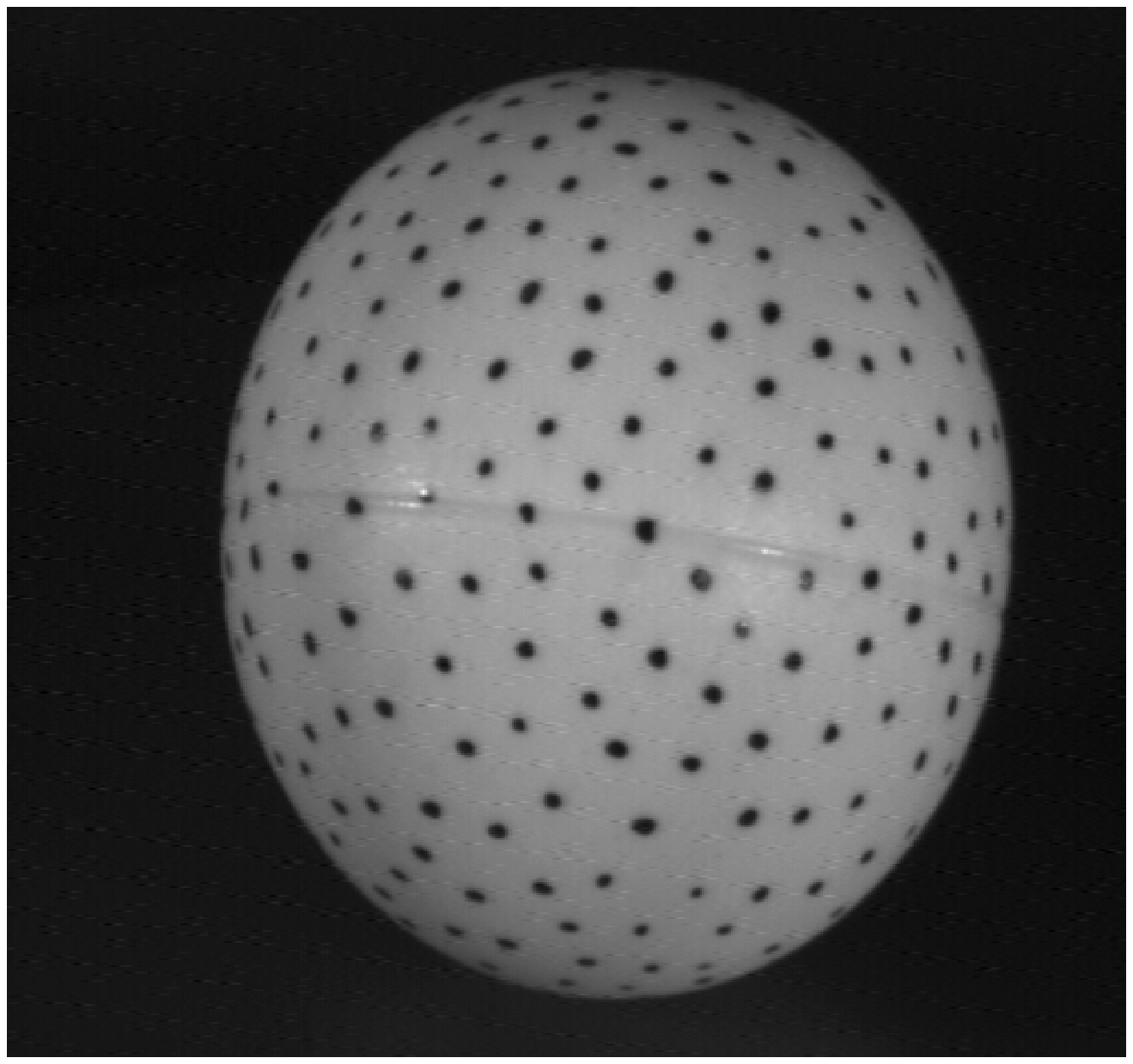,width=6cm}\hspace*{1.5cm}
 \psfig{figure=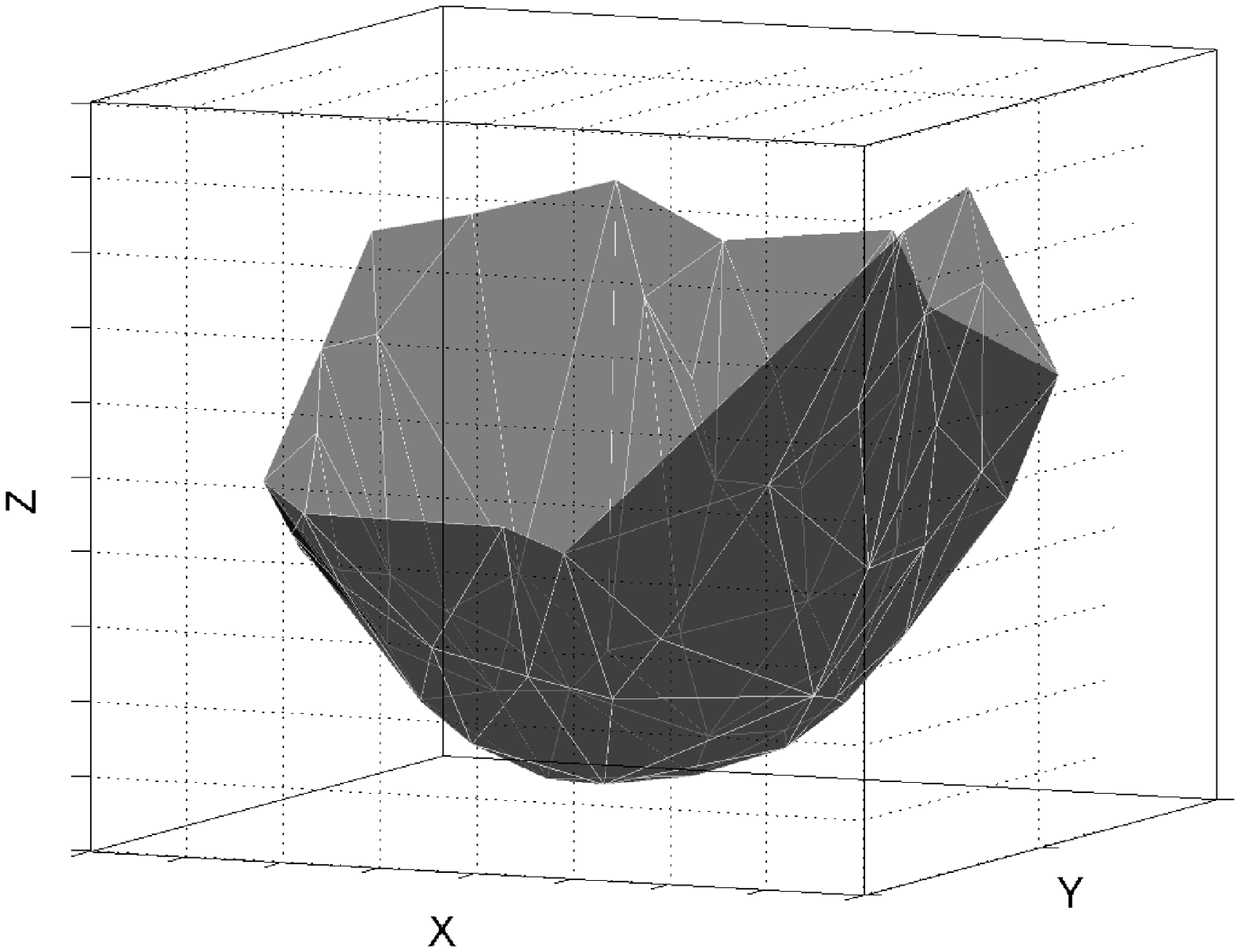,width=7.5cm}}
  \caption{Ping-pong ball video sequence. Left: first frame.
  Right: recovered 3-D~shape.\label{fig:exp3im}}
\end{figure}

We used a real-life video clip available at~\cite{cmucvsite}. This
clip shows a rotating ping-pong ball with painted dots. The left
image of Fig.~\ref{fig:exp3im} shows the first of the $52$~video
frames of the ball sequence. Due to the rotating motion, the ball
occludes itself. However, it does not complete a turn, thus there
are not re-appearing regions.

We used simple correlation techniques to track a set of
$64$~feature points (these techniques are adequate for the short
baseline video sequences we used but more sophisticated methods to
cope with wide baselines are available in the
literature~\cite{roy-chowdhury04}). Due to the camera-ball
3D~rotation, the region of the ball that is visible changes along
time, leading to an observation matrix with~$\simeq41\%$ missing
entries. We used the RC~algorithm to estimate the rank~4 matrix
that best matched the observation and computed the 3-D structure
by proceeding as in the factorization method of~\cite{tomasi92}.

In the right image of Fig.~\ref{fig:exp3im}, we represent the
recovered 3-D shape, which shows that our method succeeded in
recovering the spherical surface of the ball.

\subsection{Complete 3-D models}

%
We now illustrate our method to recover complete 3-D models from
video by minimizing a global cost function as described in
section~\ref{sec:pl}. We processed the observation matrix that
lead to the left plot of Fig.~\ref{fig:3-Dshape_wrong}. The 3-D
shape recovered by our global method is shown in the right plot,
where each of the pairs of re-appearing features have been
correctly detected as representing the same 3-D point.



\begin{figure} [htb]
 \centerline{\psfig{figure=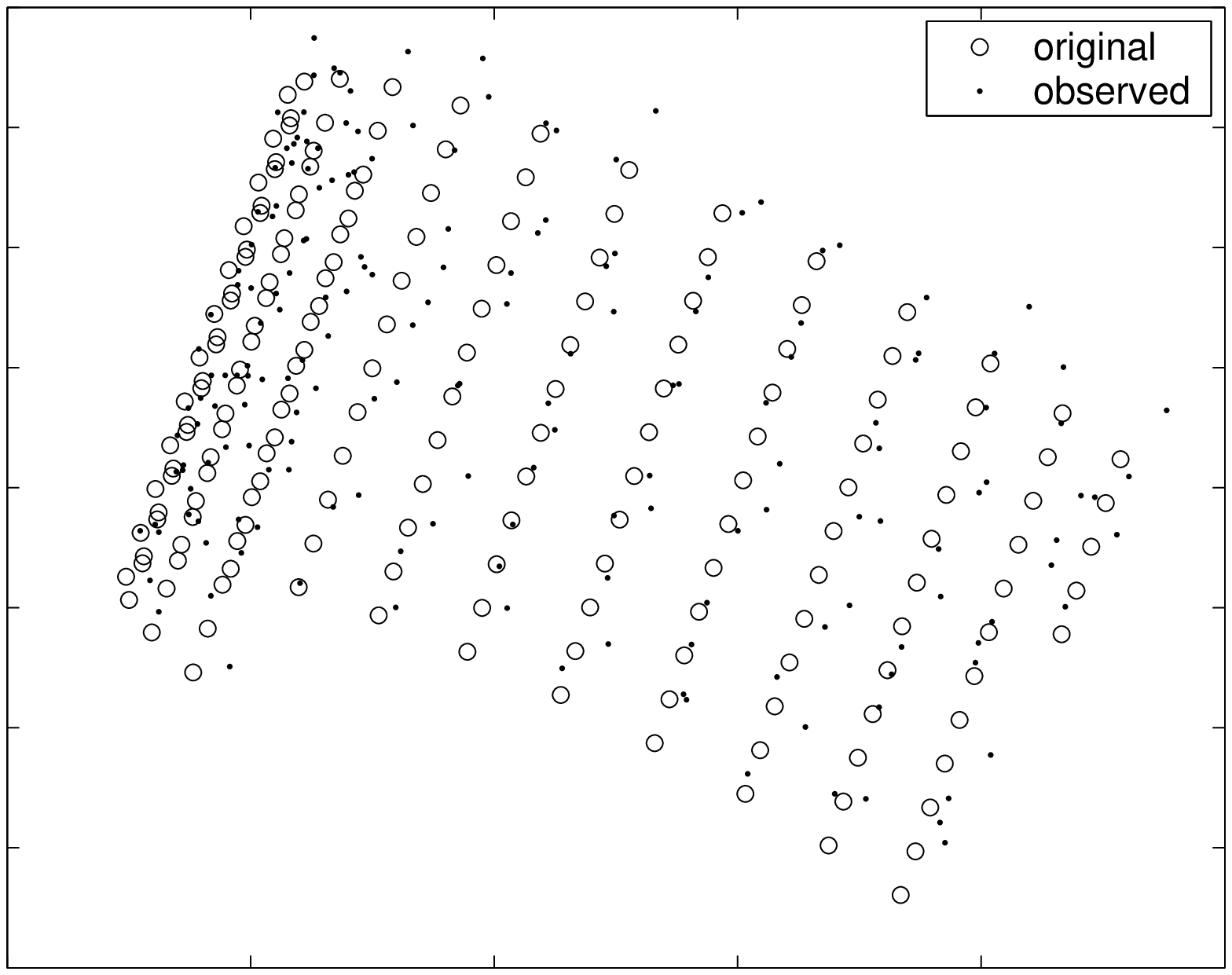,width=7cm}\hspace*{1cm}
 \psfig{figure=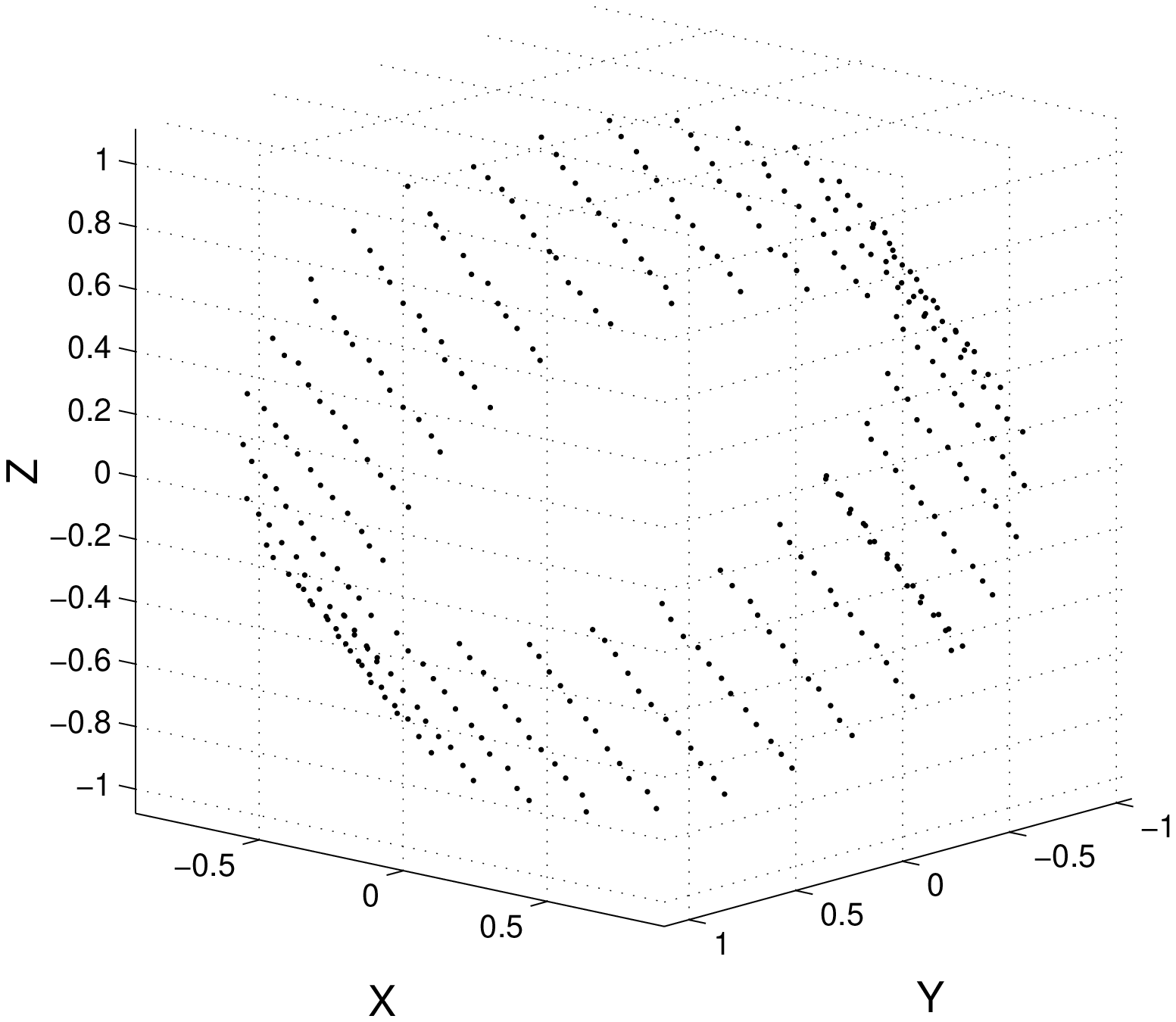,width=7.5cm}}
  \caption{Cylinder image sequence. Left: one synthetic frame. Right: recovered 3-D shape.
  \label{fig:cilim}}
\end{figure}

Fig.~\ref{fig:cilim} describes an experiment with an object
described by a larger number of 3-D feature points. We synthesized
noisy versions of the 2-D~trajectories of $372$~feature points
located on the 3-D~surface of a rotating cylinder. Then, we
simulated occlusion and inclusion by removing significant segments
of those trajectories. The left plot of Fig.~\ref{fig:cilim} shows
one of the $50$~synthesized frames. The small circles denote the
noiseless projections and the points denote their noisy version,
{\it i.e.}, the data that is observed. Note that only an
incomplete view portion of the cylinder is observed in each frame.

The data from the cylinder sequence was then collected on
a~$100\times 372$ observation matrix $\bW$ with 9537 unknown
entries ($\simeq26\%$ of the total number). We used the global
factorization algorithm to find the re-arranged matrix~$\bW_R$,
{\it i.e.}, to recover simplest 3-D rigid structure that could
have generated the data. The right plot of Fig.~\ref{fig:cilim}
plots the final estimate of the 3D~shape. We see that the complete
cylinder is recovered. Due to the incorporation of the rigidity
constraint, the 3D~positions of the features points are accurately
estimated, even in the presence of very noisy observations
(compare the plots in Fig.~\ref{fig:cilim}).

\subsection{Error analysis}

\begin{figure}[hbt]
\centerline{\psfig{figure=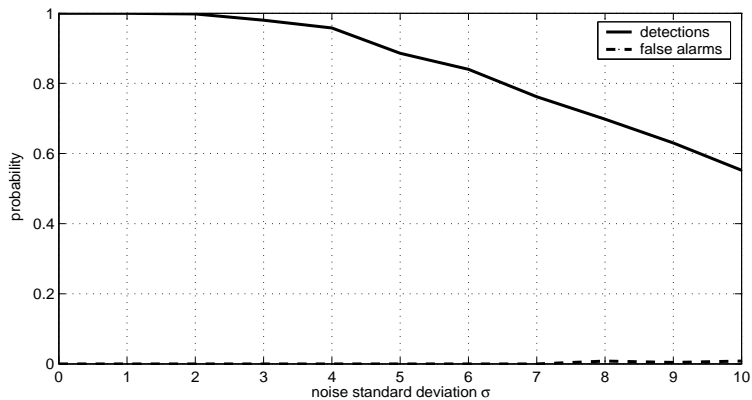,width=10cm}}
\caption{Probability of detection and probability of false
alarm.\label{fig:noise}}
\end{figure}

We quantified the gain of using our method to recover SFM by
measuring the 3-D reconstruction error. We synthesized noisy
trajectories of 40 features randomly located in 3-D, 15 of them
being ``interrupted" to simulate object self-occlusion, leading to
a $60\!\times\!55$ observation matrix~$\bW$ with 53.9\% known
entries ($x-$ and $y-$ coordinates in $[0,120]\!\times\![0,160]$,
noise standard deviation $\sigma\!=\!3$). Our method generated a
$60\times40$ re-arranged matrix~$\bW_R$ with 74.2\% known entries.
By using~$\bW_R$ rather then~$\bW$ to recover SFM, we reduced the
3-D~shape estimation error by approximately~50\% and the 3-D
motion error by approximately~70\%.
In order to evaluate the sensitivity of our method to the
observation noise, we plotted in Fig.~\ref{fig:noise} the
probabilities of correctly detecting re-appearing features and the
probability of incorrect detections (false alarms) as functions of
the noise level, for the experimental setup described above.
These probabilities were estimated from 100 runs for each level of
noise. Although re-appearing features become harder to detect as
the noise level increases, our method correctly detects more than
90\% of them when the noise level is below~$\sigma=5$. The plot of
Fig.~\ref{fig:noise} also shows that our method has approximately
zero false alarms for noise levels below~$\sigma=7$. Note that the
$x-$ and $y-$ coordinates are in $[0,120]\!\times\![0,160]$, so
this level of performance seems to indicate that our method is
able to cope with the noise in trajectories likely to happen when
processing the majority of real videos.
%

\subsection{Real video---Rubik's cube}

\begin{figure} [htb]
\centerline{{\psfig{figure=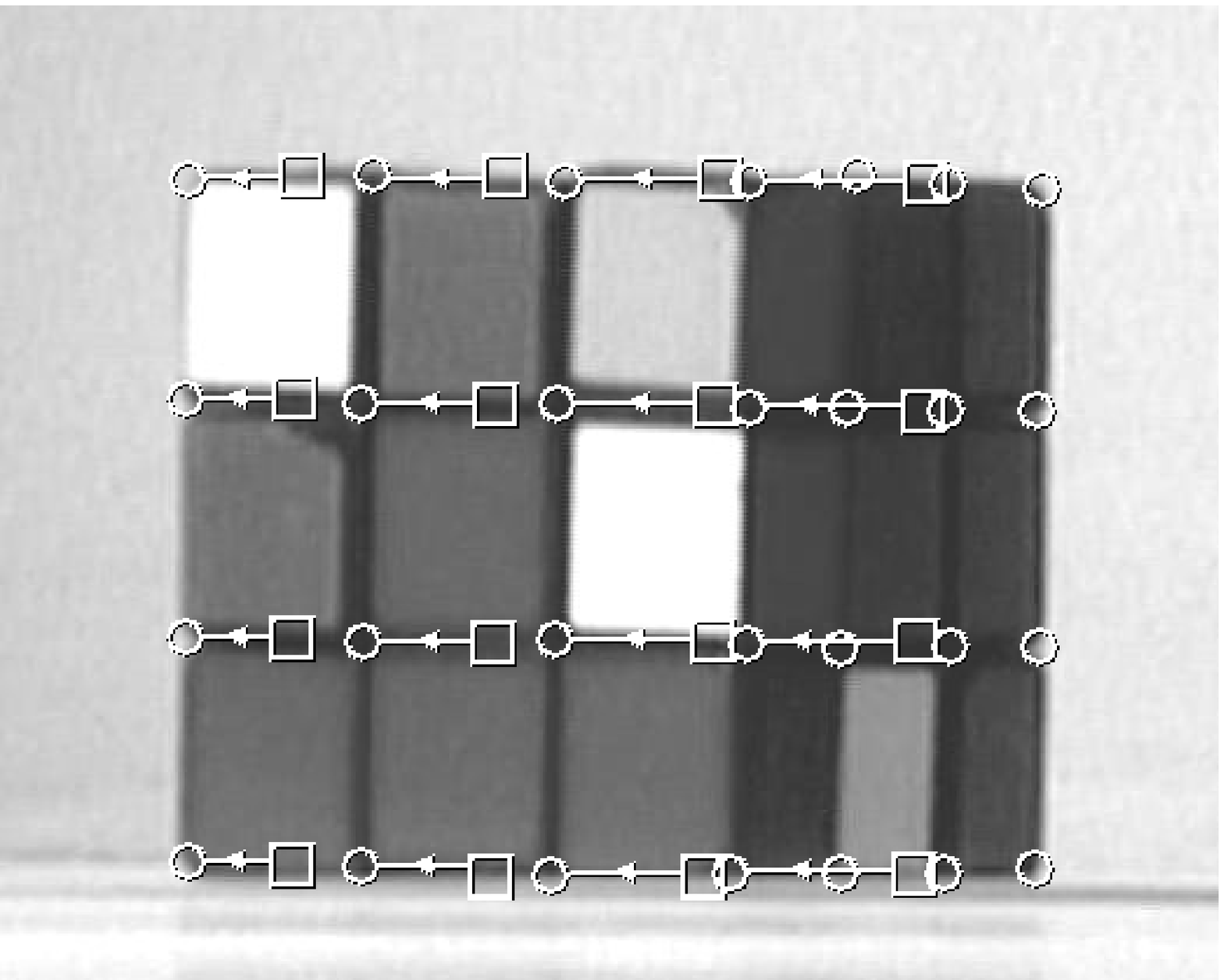,height=3.5cm}\hspace*{.5cm}
\psfig{figure=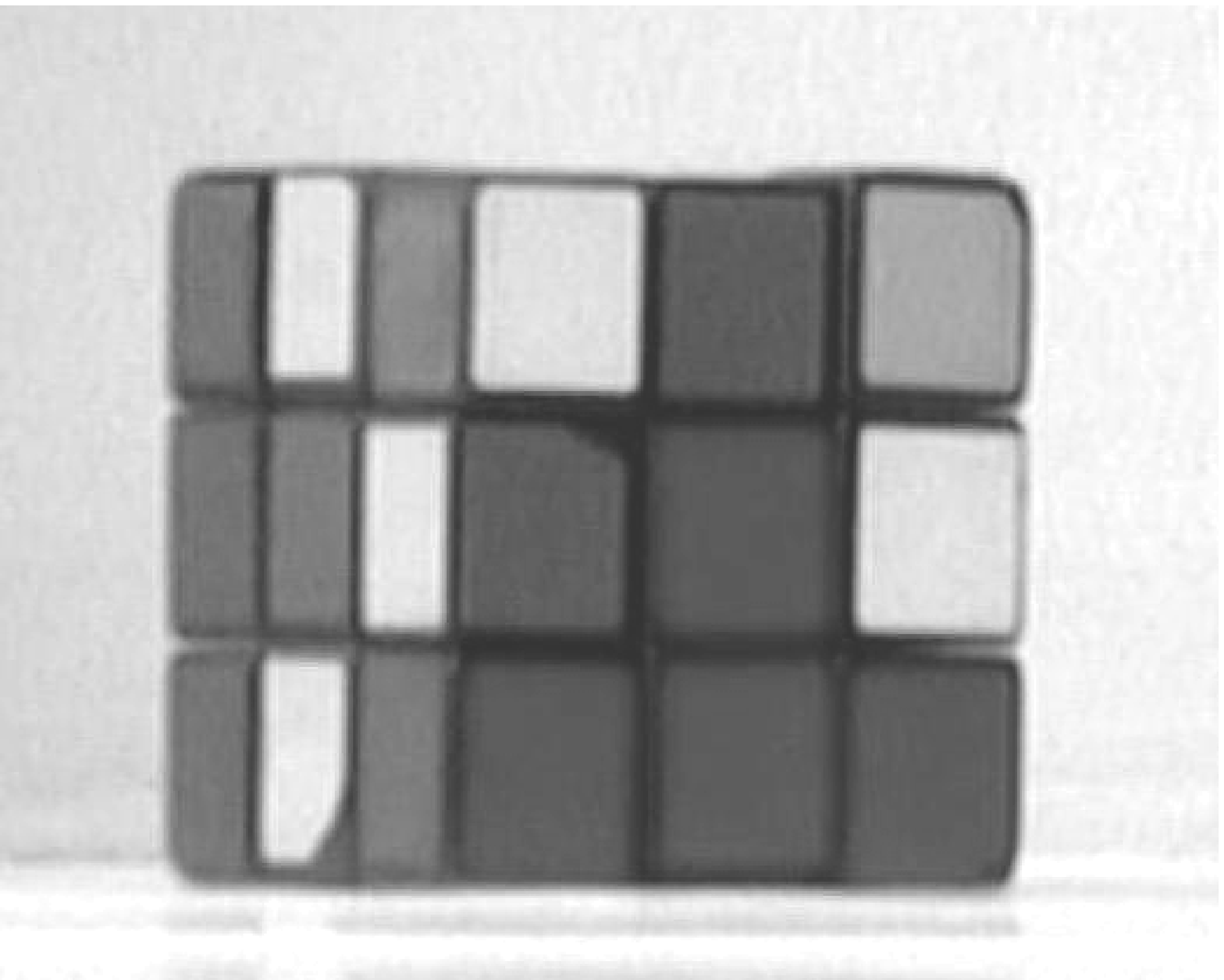,height=3.5cm}}\hspace*{1cm}
\framebox{\psfig{figure=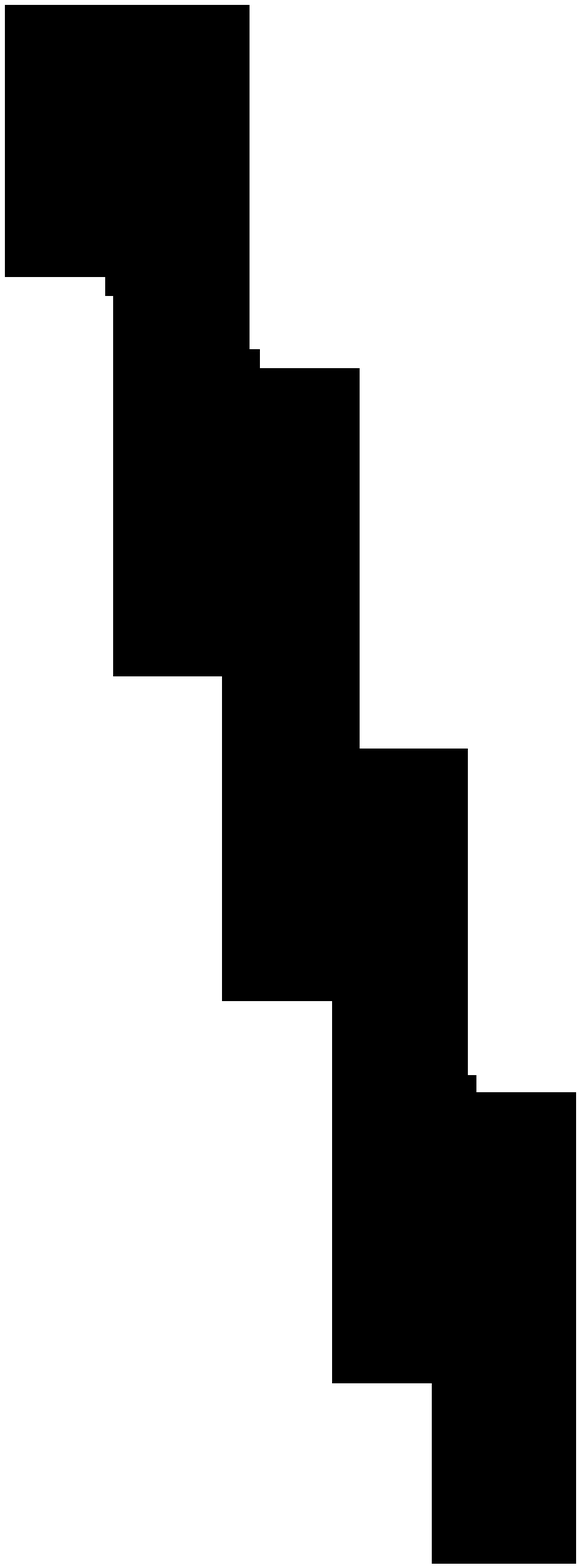,width=3.5cm,height=3.25cm}}}
\caption{Rubik's cube video clip. Left: frame with visible
features and corresponding partial trajectories. Middle: another
frame. Right: binary mask~$\bM$ representing the known entries of
the observation matrix~$\bW$---black regions correspond to entries
$m_{ij}=1$ meaning that $w_{ij}$ is observed, {\em i.e.},
feature~$j$ is visible in frame~$i$; white regions represent the
opposite.
  \label{fig:m_cubes}}
\end{figure}

This video---see two representative frames on the left and middle
images of Fig.~\ref{fig:m_cubes}---shows a Rubik's cube. The video
sequence was obtained by rotating a hand-held camera around a
table with the cube. In the leftmost image of
Fig.~\ref{fig:m_cubes}, we superimposed with the video frame the
visible features and the initial parts of their trajectories. Due
to the cube self-occlusion, feature points enter and leave the
scene. In particular, along the video sequence, all features
disappear and several of them re-appear because the camera
performs a complete turn around the cube.

\begin{figure} [htb]
 \centerline{\psfig{figure=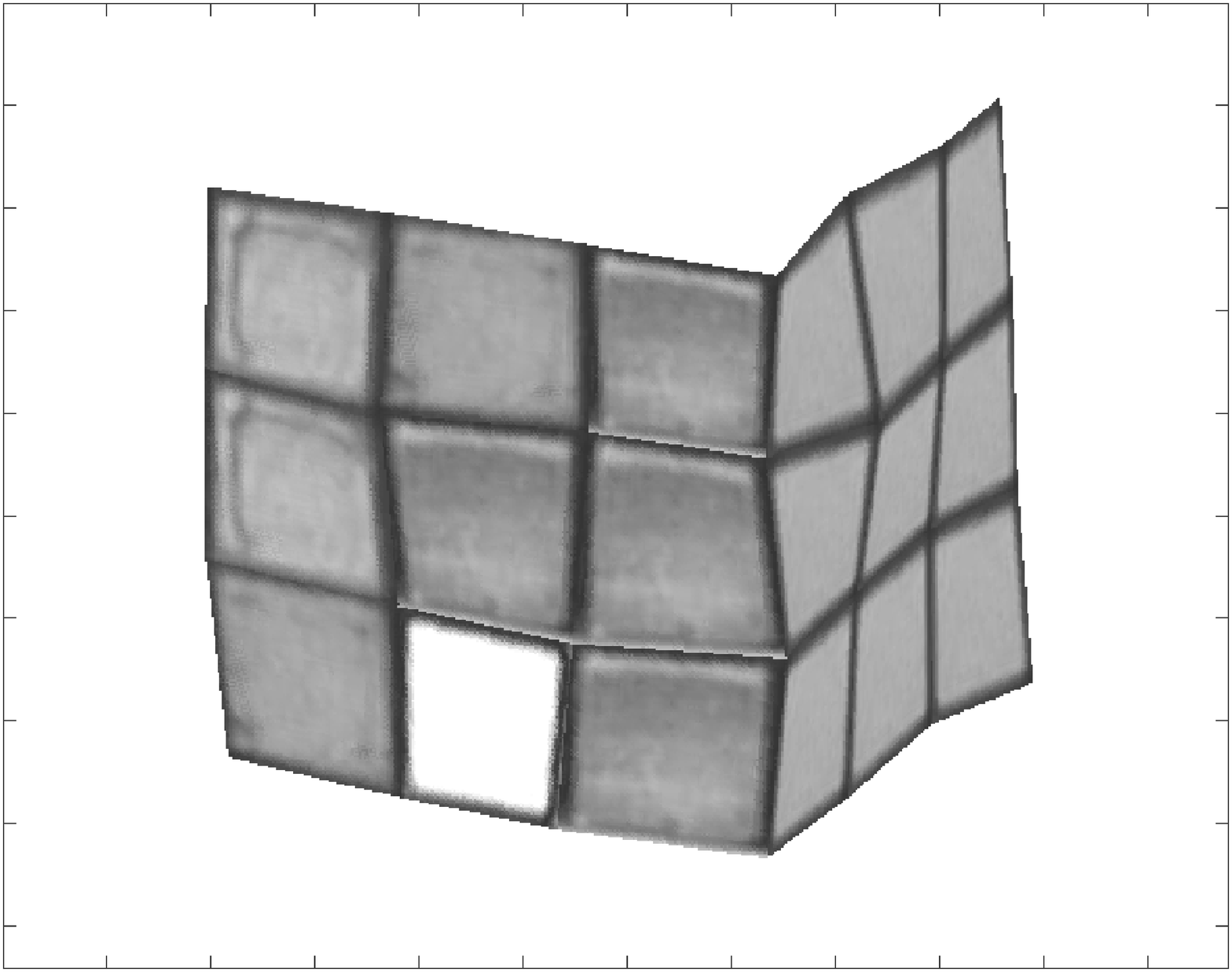,width=7cm}\hspace*{1cm}
 \psfig{figure=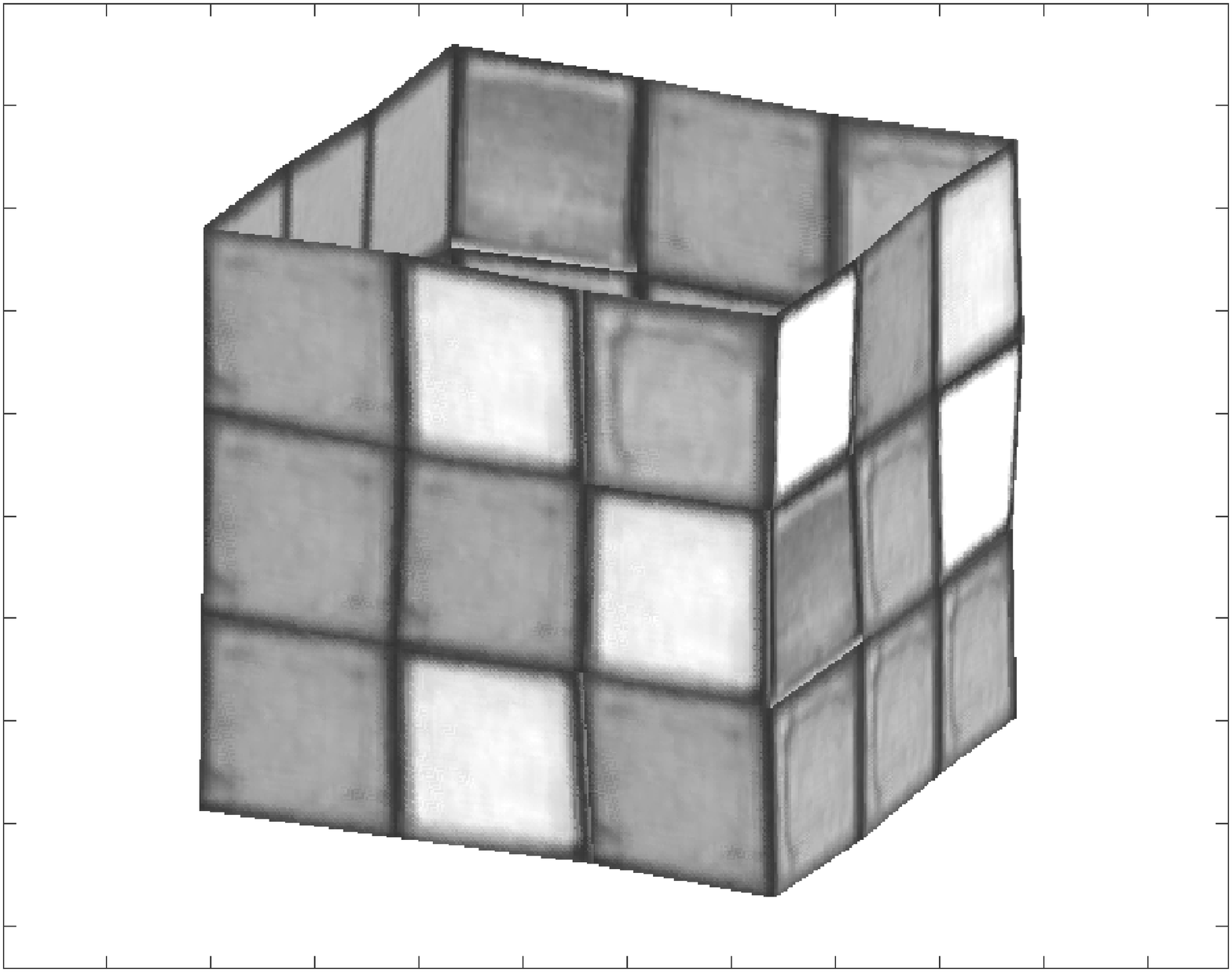,width=7cm}}
\caption{Texture mapped 3D shape recovered from the Rubik's cube
video clip in Fig.~\ref{fig:m_cubes}. Left: incomplete model
obtained by using the factorization method of Tomasi and
Kanade~[1]. Right: complete shape recovered by our global
factorization.\label{fig:3D_cube}}
\end{figure}

To emphasize the advantages of using the algorithms described in
this paper, we first applied to a segment of the Rubik's cube
video clip the original factorization method of~\cite{tomasi92}
for complete matrices, obtaining the 3D~shape represented on the
left side of Fig.~\ref{fig:3D_cube}. This model was obtained with
$28$~features and $18$~frames.

We then collected the entire set of the visible parts of the
trajectories of $64$~features across $85$~frames in a
$170\!\times\! 64$~incomplete observation matrix~$\bW$. The
structure of the missing part of~$\bW$ is coded by the $170\times
64$~binary mask~$\bM$ represented on the right side of
Figure~\ref{fig:m_cubes}. The number of missing entries in~$\bW$
was about~$62\%$.
We computed the re-arranged matrix~$\bW_R$ by using our global
factorization. The right image of Fig.~\ref{fig:3D_cube} shows the
texture-mapped 3-D shape recovered from~$\bW_R$.

This example illustrates how our global approach trivializes the
usually hard task of merging partial estimates of a 3D~model like
the one in the left image of Fig.~\ref{fig:3D_cube}. Note that the
top face is missing also in the right image of
Fig.~\ref{fig:3D_cube} because the position the cube model is
shown in that image was not seen in the video sequence. The
advantage of using our approach to recover 3-D~rigid structure is
two-fold. First, while recovering a complete 3-D~model by fusing
partial models as the one on the left side of
Fig.~\ref{fig:3D_cube} is a complex task, our method recovers
directly the complete model shown on the right side of
Fig.~\ref{fig:3D_cube}. Second, rather than processing subsets of
the set of features and frames at disjoint steps, our method uses
all the information available in a global way, leading to more
accurate 3-D~shapes as illustrated by the 3-D~models in
Fig.~\ref{fig:3D_cube}.

\subsection{Video compression}
The 3D~models recovered by our method can be used to represent in
an efficient way the original video sequence as proposed
in~\cite{aguiar99-3}---the video sequence is represented by the
3-D~shape, texture, and 3-D~motion of the objects. This leads to
significant bandwidth saving since once the 3-D~shape and texture
of the objects have been transmitted, their 3-D~motion is
transmitted with a few bytes per frame.

\begin{figure} [htb]
\centerline{\psfig{figure=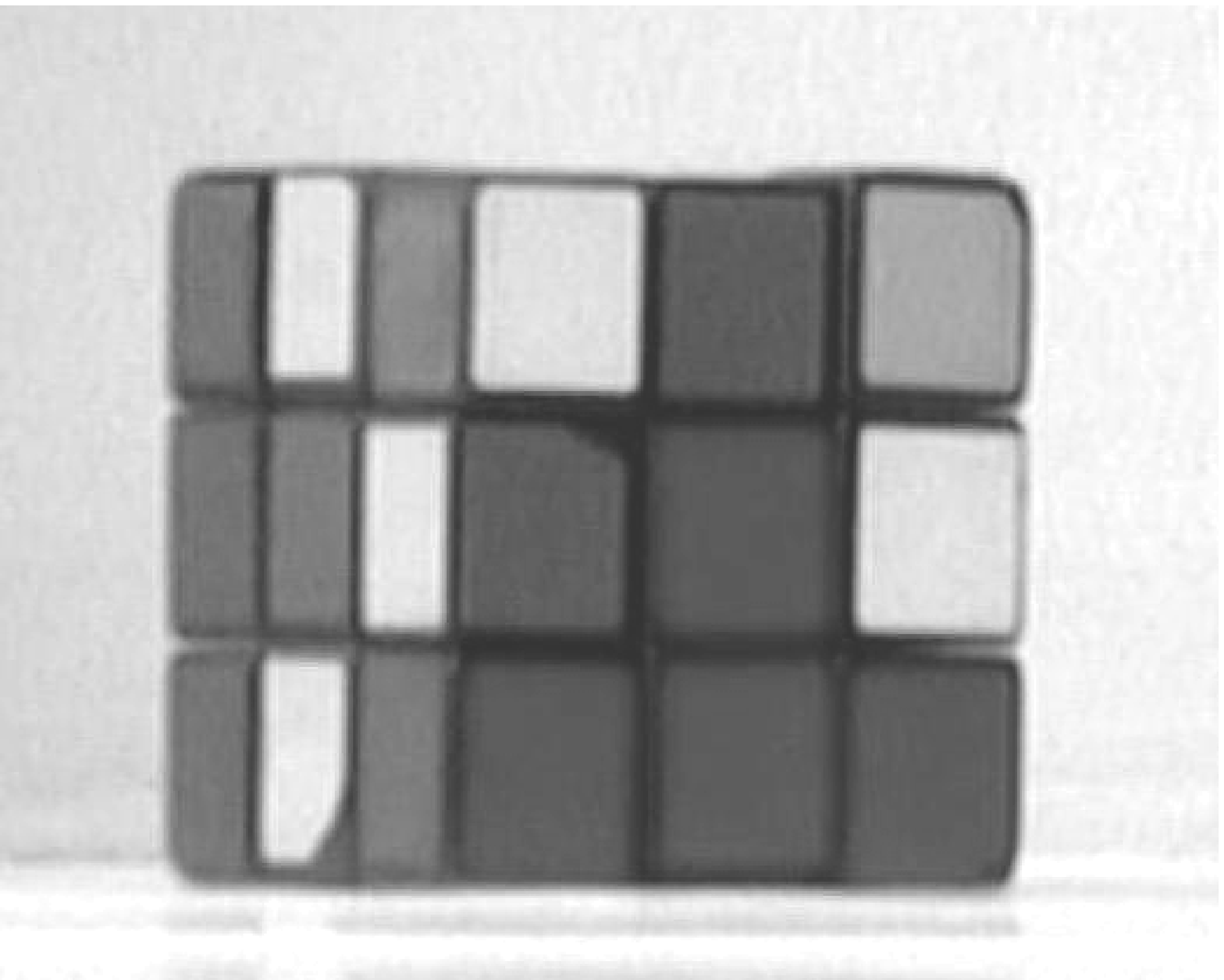,width=3.5cm}\hspace*{.1cm}\psfig{figure=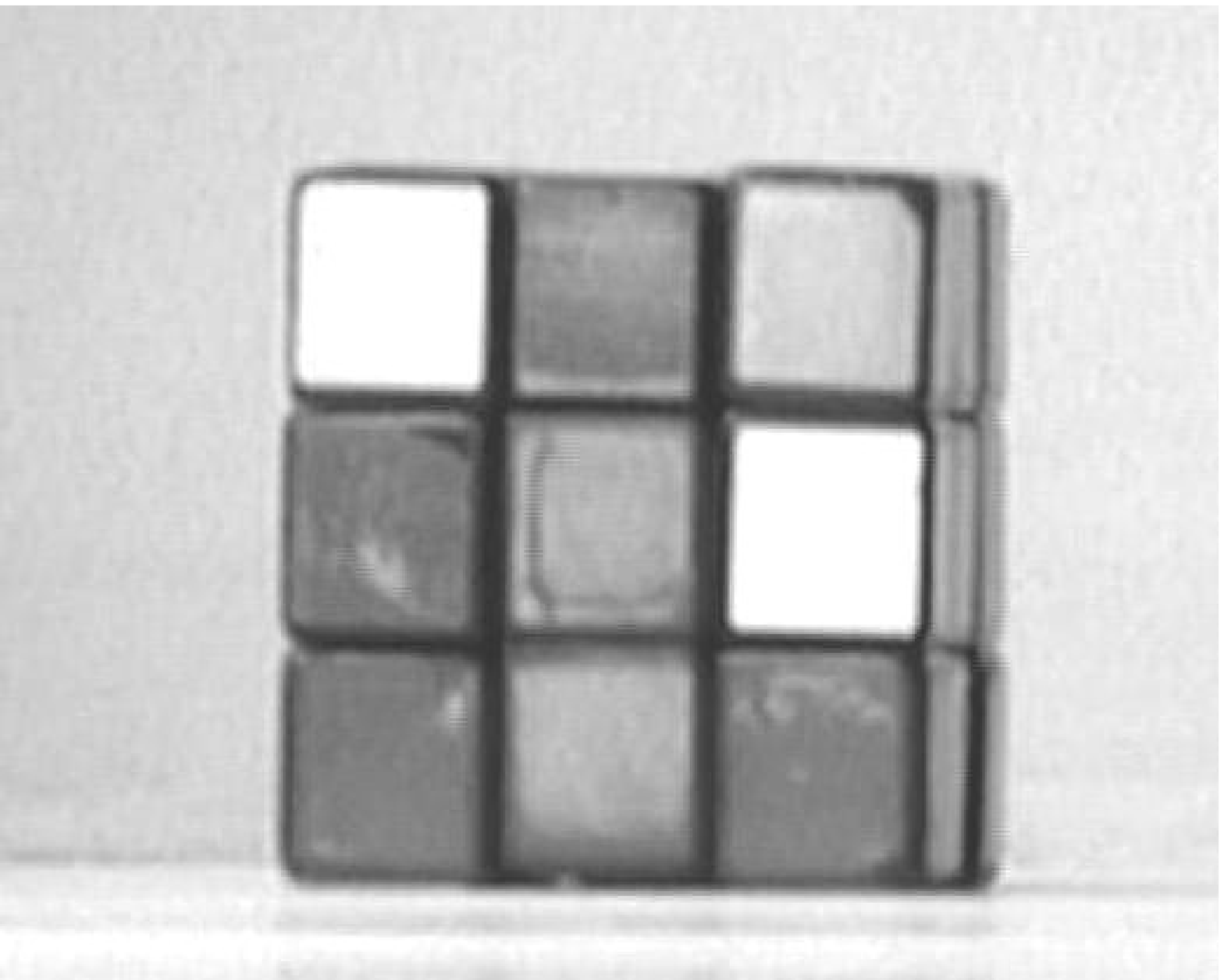,width=3.5cm}\hspace*{.1cm}\psfig{figure=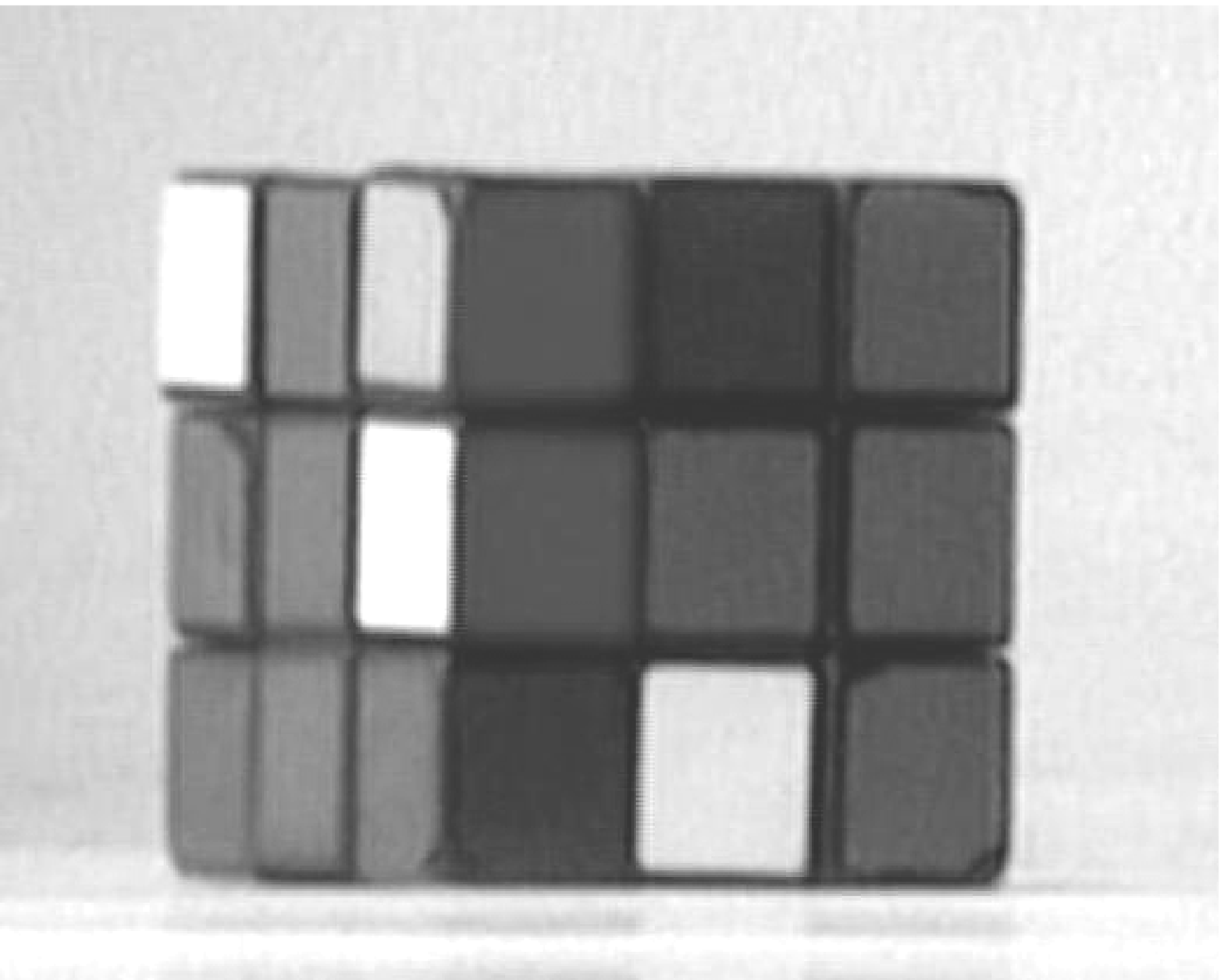,width=3.5cm}\hspace*{.1cm}\psfig{figure=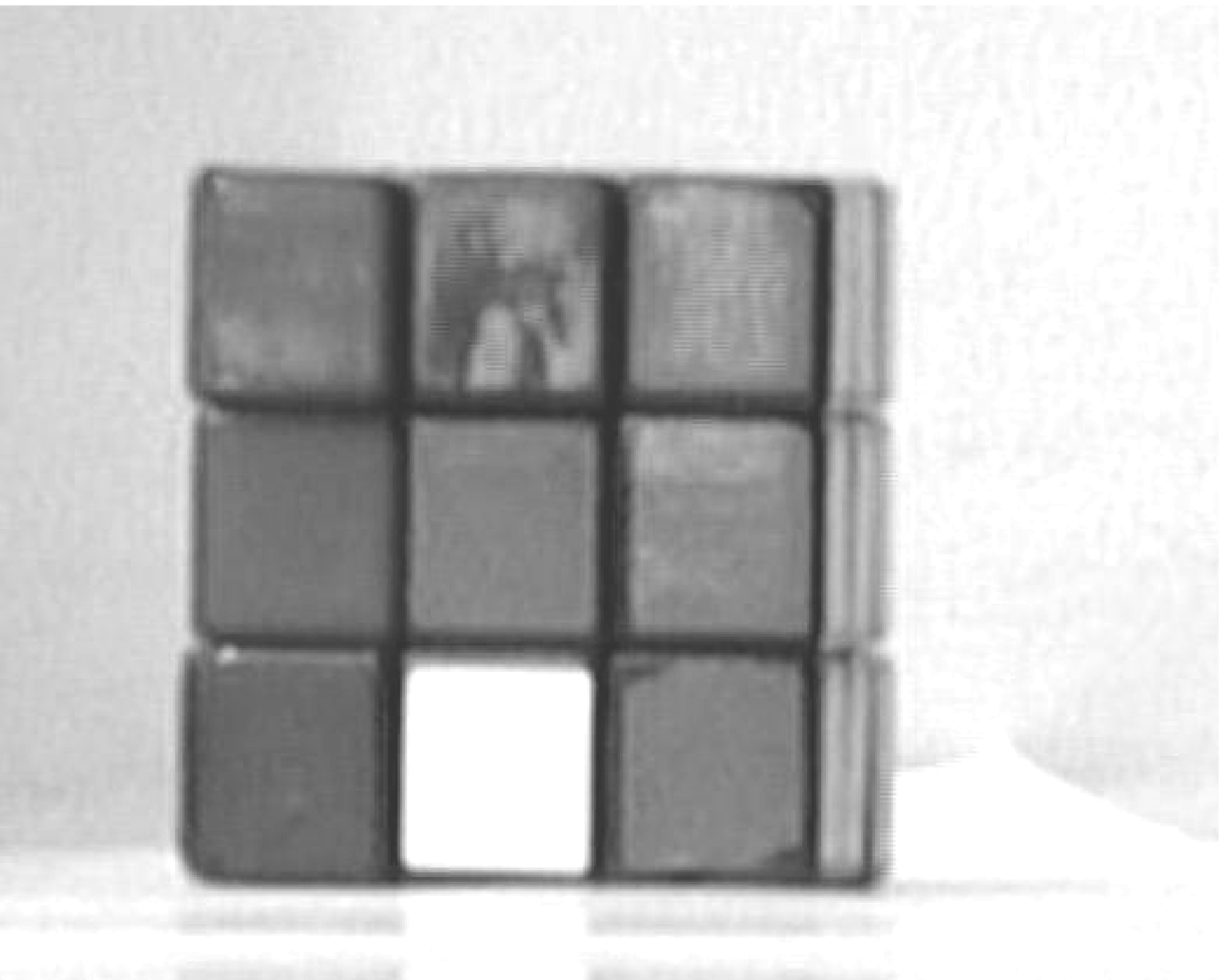,width=3.5cm}}
\vspace*{.1cm}
\centerline{\psfig{figure=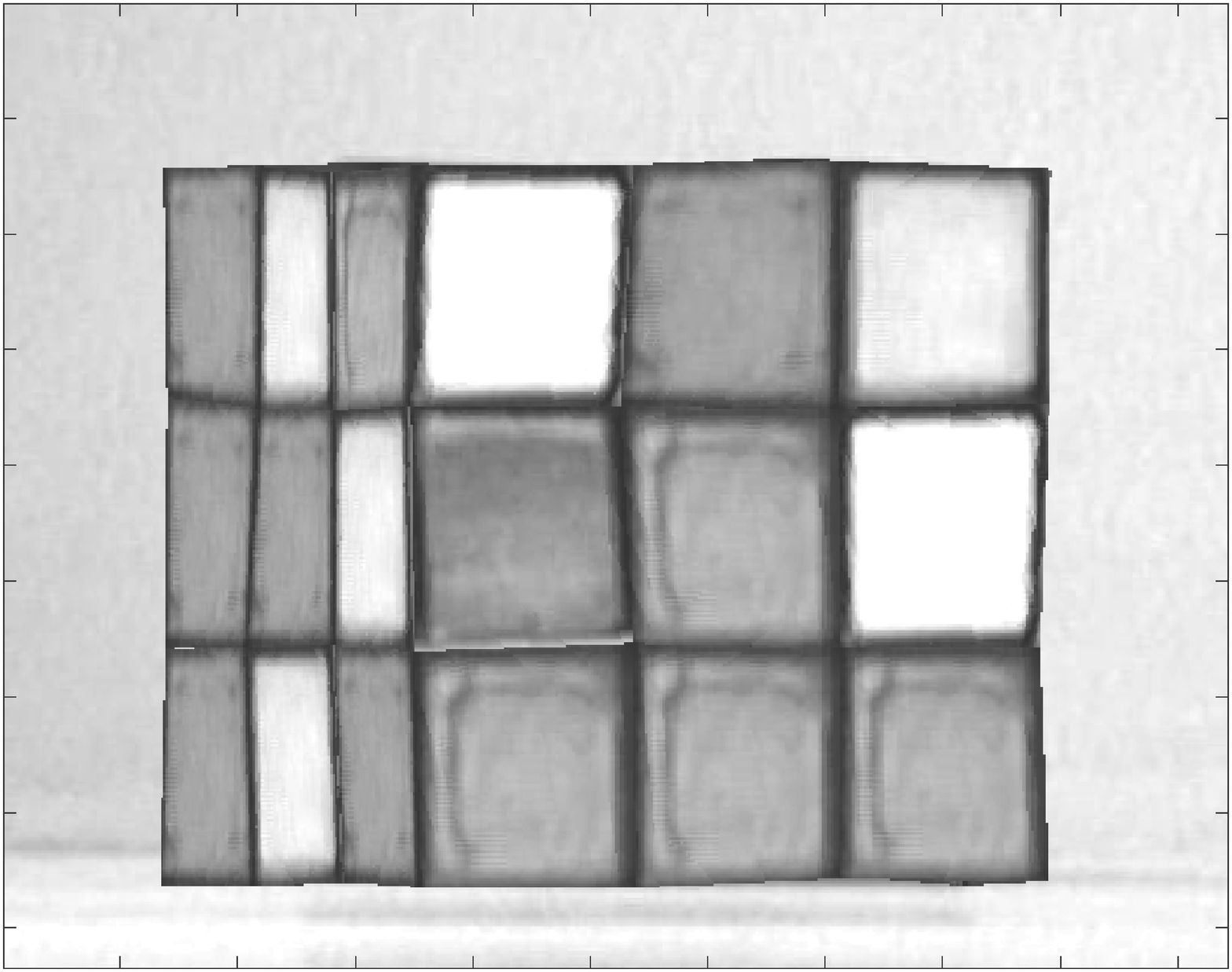,width=3.5cm}\hspace*{.1cm}\psfig{figure=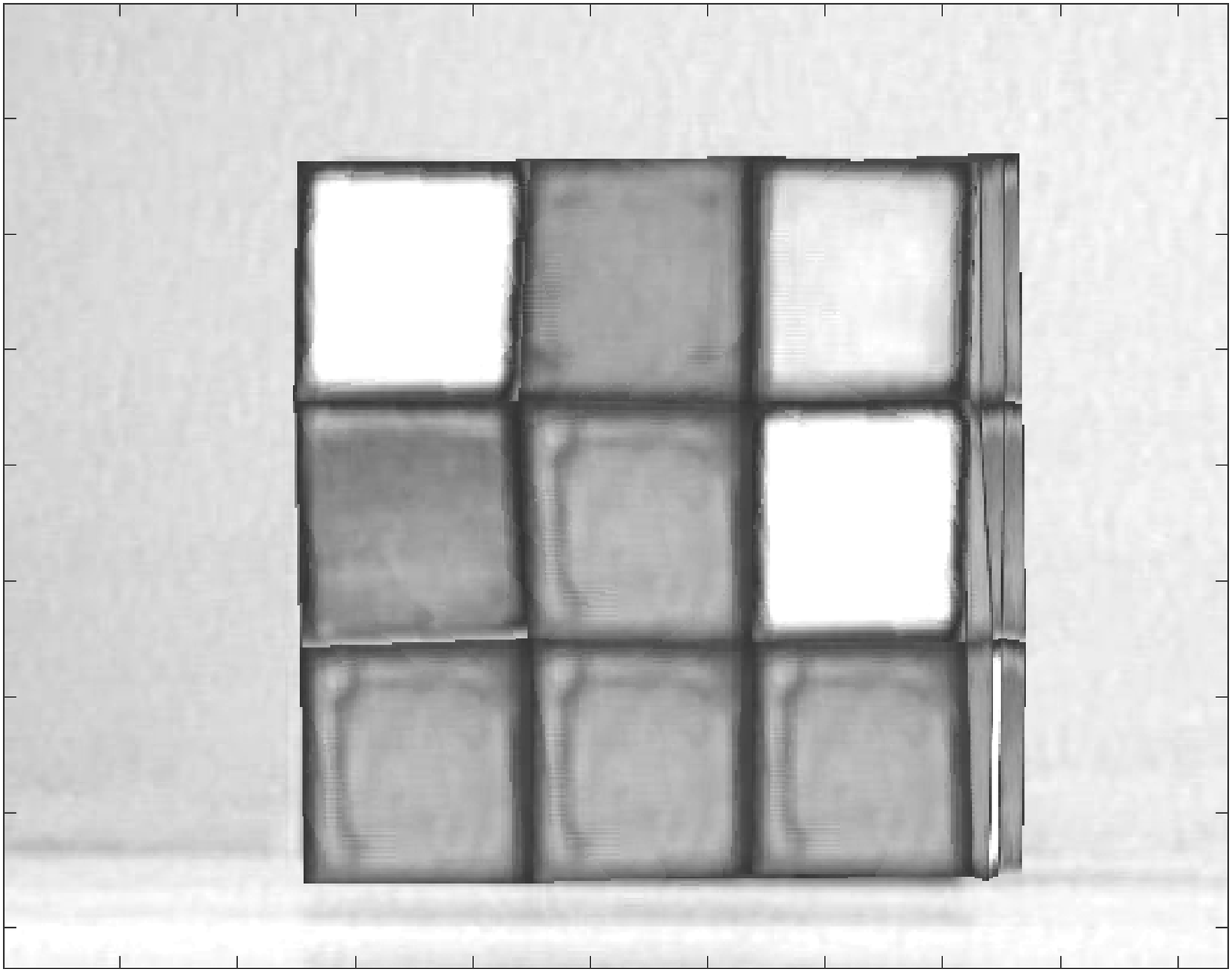,width=3.5cm}\hspace*{.1cm}\psfig{figure=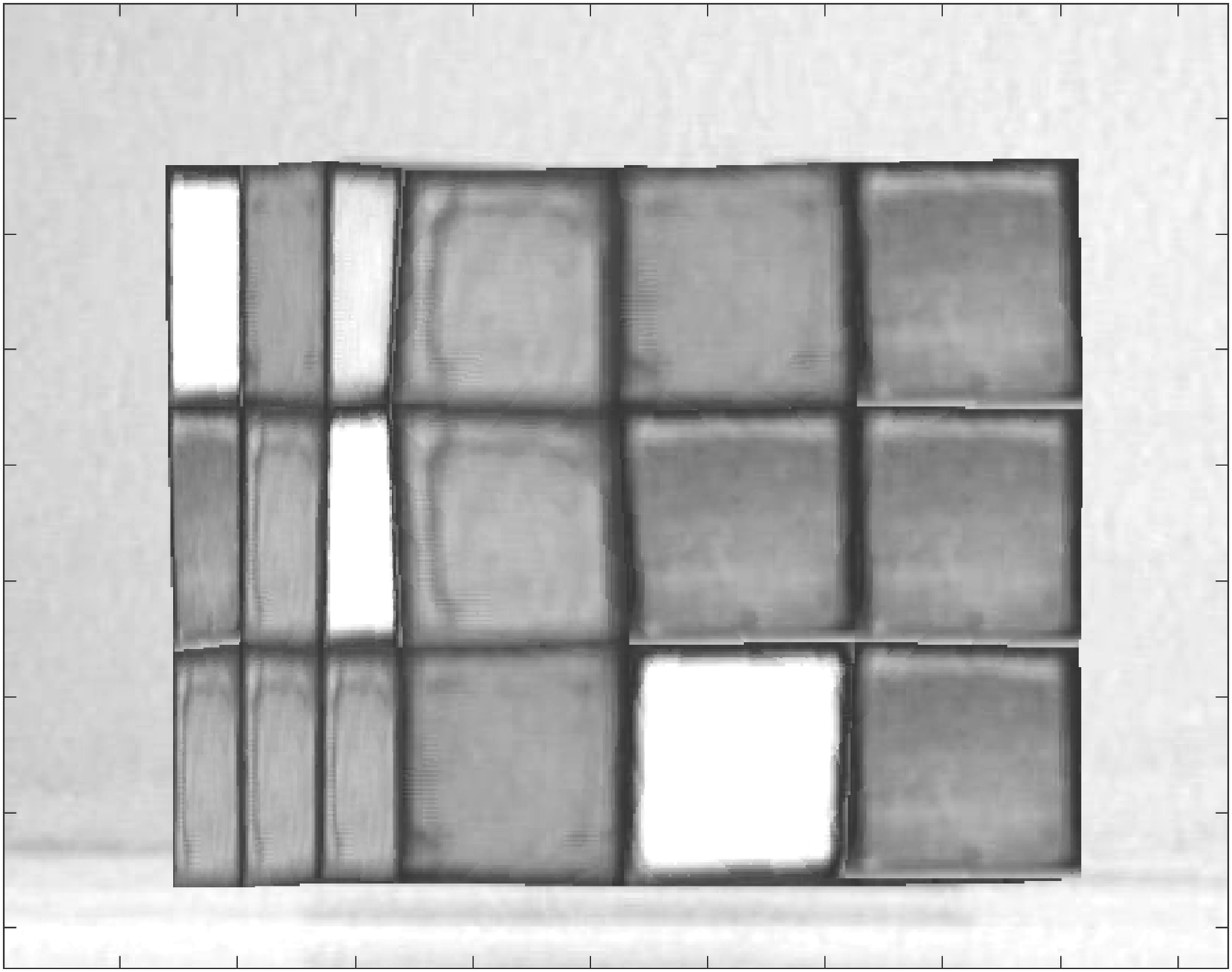,width=3.5cm}\hspace*{.1cm}\psfig{figure=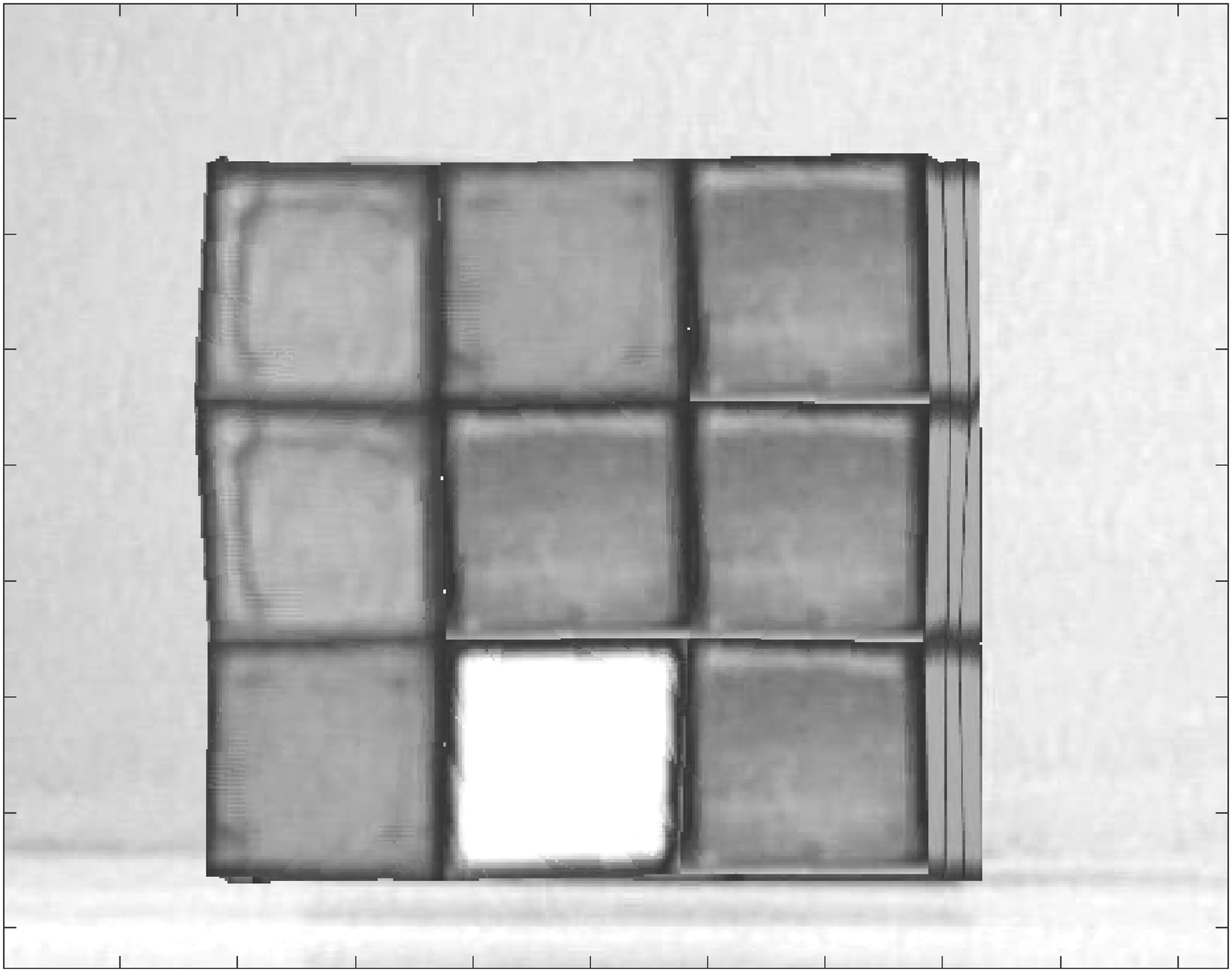,width=3.5cm}}
\caption{3-D~model-based digital video compression
example. Top row: original frames. Bottom row: compressed frames.
Compression ratio approx. $10^3$.
  \label{fig:compression}}
\end{figure}

We used this methodology to compress the entire sequence of
$2161$~frames of the Rubik's cube video clip. The compression
ratio (relative to the original J-PEG compressed frames) was
approximately~$10^3$. Fig.~\ref{fig:compression} shows sample
original frames (top row) and the corresponding compressed frames
(bottom row). The differences of lighting between the top and
bottom images are due to fact that the constancy of the texture of
the 3-D~model does not model the different light reflections that
happen in the real scenario.


%

\section{Conclusion}\label{sec:conc}

We proposed a global approach to build complete 3-D models from
video. The 3-D model is inferred as the {\it simplest} rigid
object that agrees with all the observed data, {\it i.e.}, with
the {\it entire} set of video frames. Since a key step in this
global approach is the approximation of rank deficient matrices
from incomplete observations, we developed two iterative
algorithms to solve this non-linear problem: {\it
Expectation-Maximization}~(EM) and {\it Row-Column}~(RC). Our
experiments showed that both algorithms converged to the correct
estimate whenever initialized by a simple procedure and that the
RC algorithm is computationally cheaper and more robust than~EM.
Our experiments show that the proposed global method leads to more
accurate estimates of 3-D~shape than those obtained by either
processing several smaller subsets of views or processing the
entire video without taking into account re-appearing regions.

\appendix


This appendix addresses the computation of a rank~4 initial
guess~$\widehat{\bW_r}^{(0)}$ from an incomplete
observation~$\bW_r$. We find~$\widehat{\bW_r}^{(0)}$ by computing,
in an expedite way, guesses of its column space matrix~$\bA$ and
its row space matrix~$\bB$, from the data in~$\bW_r$:
\begin{equation}\label{eq:ab2}
  \widehat{\bW_r}^{(0)}=\bA_{2F\times 4}\bB_{4\times P_r}\in\cS_4 .
\end{equation}

\subsection{Subspace approximation}

Before addressing the general case, we consider the simpler case
where a number of columns of~$\bW_r$ are entirely known, {\it
i.e.}, a number of feature points are present in all frames, and a
number of rows of~$\bW_r$ are entirely known, {\it i.e.}, and a
number of frames contain all feature projections. We collect those
known columns in a submatrix~$\bW_A$ and those known rows in a
submatrix~$\bW_B$. From the data in~$\bW_A$, we estimate the
column space matrix~$\bA$ as
\begin{equation}\label{eq:a}
  \bA=\bW_A\downarrow\cS_4.
\end{equation}
From~$\bW_B$ and the column space~$\bA$, the LS estimate of the
row space matrix~$\bB$ is, see~\cite{golub96},
\begin{equation}\label{eq:b}
  \bB=\left(\bA_B^T\bA_B\right)^{-1}\bA_B^T\bW_B,
\end{equation}
where~$\bA_B$ collects the rows of~$\bA$ that correspond to the
rows of~$\bW_B$. We see that the matrix~$\bA_B^T\bA_B$ must be
nonsingular so the matrix~$\bW_A$ must have at least 4 linearly
independent columns and the matrix~$\bW_B$ must have at least 4
linearly independent rows.
%

\subsection{Subspace combination}

In general, it may be impossible to find 4~entire columns and
4~entire rows without missing elements in the matrix~$\bW_r$. We
must then estimate the column and row space matrices~$\bA$
and~$\bB$ by combining the spaces that correspond to smaller
submatrices of~$\bW_r$. We describe the algorithm for combining
two column/row space matrices. The process is then repeated until
the entire matrix~$\bW_r$ has been processed.

Select from~$\bW$ two submatrices~$\bW_1$ and~$\bW_2$ that have at
least 4~columns and 4~rows without missing elements. We
factorize~$\bW_1$ and~$\bW_2$ using~(\ref{eq:a}) to obtain the
corresponding column space matrices~$\bA_1$ and~$\bA_2$. If the
observation matrix~$\bW$ is in fact well modelled by a rank~4
matrix, the submatrices of~$\bA_1$ and~$\bA_2$ that correspond to
common rows in~$\bW$, denoted by~$\bA_{12}$ and~$\bA_{21}$, are
related by a linear transformation,
\begin{equation}\label{eq:a12}
  \bA_{12}\simeq\bA_{21}\bN_{4\times 4}.
\end{equation}
We compute~$\bN$ as the LS~solution of~(\ref{eq:a12}) and assemble
a combined column space matrix~$\bA$ for the rows corresponding
to~$\bW_1$ and~$\bW_2$:
\begin{equation}\label{eq:n}
  \bN=\left(\bA_{21}^T\bA_{21}\right)^{-1}\bA_{21}^T\bA_{12},\quad\qquad\mbox{and}\quad\quad
  \bA=\left[\begin{array}{c}\bA_{1}\\\hline
  \bA_{2\setminus 1}\bN\end{array}\right],
\end{equation}
where~$\bA_{2\setminus 1}$ denotes the submatrix of~$\bA_2$ that
collects the rows that do not correspond to rows of~$\bW_1$. We
compute the combined row space matrix~$\bB$ by using an analogous
procedure.

\bibliography{global3D}

\begin{thebibliography}{10}
\providecommand{\url}[1]{#1}
\csname url@rmstyle\endcsname
\providecommand{\newblock}{\relax}
\providecommand{\bibinfo}[2]{#2}
\providecommand\BIBentrySTDinterwordspacing{\spaceskip=0pt\relax}
\providecommand\BIBentryALTinterwordstretchfactor{4}
\providecommand\BIBentryALTinterwordspacing{\spaceskip=\fontdimen2\font plus
\BIBentryALTinterwordstretchfactor\fontdimen3\font minus
  \fontdimen4\font\relax}
\providecommand\BIBforeignlanguage[2]{{%
\expandafter\ifx\csname l@#1\endcsname\relax
\typeout{** WARNING: IEEEtran.bst: No hyphenation pattern has been}%
\typeout{** loaded for the language `#1'. Using the pattern for}%
\typeout{** the default language instead.}%
\else
\language=\csname l@#1\endcsname
\fi
#2}}

\bibitem{Aanaes:etal:pami2002}
H.~Aanaes, R.~Fisker, K.~Astrom, and J.~Carstensen, ``Robust factorization,''
  \emph{IEEE Trans. on Pattern Analysis and Machine Intelligence}, vol.~24,
  no.~9, 2002.

\bibitem{aguiar99-1}
P.~Aguiar and J.~Moura, ``Factorization as a rank~1 problem,'' in \emph{Proc.
  of IEEE Int. Conf. on Computer Vision and Pattern Recognition}, Fort Collins,
  CO, USA, 1999.

\bibitem{aguiar99-3}
------, ``Fast 3{D} modelling from video,'' in \emph{Proc. of IEEE Int. W. on
  Multimedia Signal Proc.}, Copenhagen, Denmark, 1999.

\bibitem{aguiar01}
------, ``Three-dimensional modeling from two-dimensional video,'' \emph{IEEE
  Trans. on Image Proc.}, vol.~10, no.~10, 2001.

\bibitem{aguiar03}
------, ``Rank 1 weighted factorization for 3{D} structure recovery: Algorithms
  and performance analysis,'' \emph{IEEE Trans. on Pattern Analysis and Machine
  Intelligence}, vol.~25, no.~9, 2003.

\bibitem{irani02}
P.~Anandan and M.~Irani, ``Factorization with uncertainty,'' \emph{Kluwer Int.
  Journal of Computer Vision}, vol.~49, no. 2-3, 2002.

\bibitem{ayache91}
N.~Ayache, \emph{Artificial Vision for Mobile Robots}.\hskip 1em plus 0.5em
  minus 0.4em\relax Cambridge, MA, USA: The MIT Press, 1991.

\bibitem{barron98}
A.~Barron, J.~Rissanen, and B.~Yu, ``The minimum description length principle
  in coding and modeling,'' \emph{IEEE Trans. on Information Theory}, vol.~44,
  no.~6, 1998.

\bibitem{basri88}
R.~Basri and S.~Ullman, ``The alignment of objects with smooth surfaces,''
  \emph{Computer Graphics, Vision, and Image Processing: Image Understanding},
  vol.~57, no.~3, 1988.

\bibitem{belhumeur96}
P.~Belhumeur and D.~Kriegman, ``What is the set of images of an object under
  all possible lighting conditions~?'' in \emph{Proc. of IEEE Int. Conf. on
  Computer Vision and Pattern Recognition}, San Francisco, CA, USA, 1996.

\bibitem{berger93}
J.~Berger, \emph{Statistical Decision Theory and Bayesian Analysis}.\hskip 1em
  plus 0.5em minus 0.4em\relax New York: Springer-Verlag, 1993.

\bibitem{Brandt:smvp2002}
S.~Brandt, ``Closed form solutions for affine reconstruction under missing
  data,'' in \emph{Statistical Methods for Video Processing (ECCV Workshop)},
  Copenhagen, Denmark, 2002.

\bibitem{buchanan05}
A.~Buchanan and A.~Fitzgibbon, ``Damped newton algoritms for matrix
  factorization with missing data,'' in \emph{Proc. of IEEE Int. Conf. on
  Computer Vision and Pattern Recognition}, San Diego CA, USA, 2005.

\bibitem{chen04}
P.~Chen and D.~Suter, ``Recovering the missing components in a large noisy
  low-rank matrix: Application to {SFM},'' \emph{IEEE Trans. on Pattern
  Analysis and Machine Intelligence}, vol.~26, no.~8, 2004.

\bibitem{costeira98}
J.~Costeira and T.~Kanade, ``A factorization method for independently moving
  objects,'' \emph{Kluwer Int. Journal of Computer Vision}, vol.~29, no.~3,
  1998.

\bibitem{golub96}
G.~Golub and C.~V. Loan, \emph{Matrix Computations}.\hskip 1em plus 0.5em minus
  0.4em\relax The Johns Hopkins University Press, 1996.

\bibitem{goncalves04}
B.~Gon\c{c}alves and P.~Aguiar, ``Complete 3-{D} models from video: a global
  approach,'' in \emph{Proc. of IEEE Int. Conf. on Image Processing},
  Singapore, 2004.

\bibitem{green98penalized}
P.~Green, ``Penalized likelihood,'' in \emph{Encyclopedia of Statistical
  Sciences}.\hskip 1em plus 0.5em minus 0.4em\relax New York: John Wiley \&
  Sons, 1998.

\bibitem{guerreiro02}
R.~Guerreiro and P.~Aguiar, ``3{D} structure from video streams with partially
  overlapping images,'' in \emph{Proc. of IEEE Int. Conf. on Image Processing},
  New York, USA, 2002.

\bibitem{guerreiro02a}
------, ``Factorization with missing data for 3d structure recovery,'' in
  \emph{Proc. of IEEE Int. Workshop on Multimedia Signal Processing}, St.
  Thomas, USA, 2002.

\bibitem{guerreiro03}
------, ``Estimation of rank deficient matrices from partial observations:
  two-step iterative algorithms,'' in \emph{Energy Min. Meth. in Computer
  Vision and Pattern Recognition}, ser. Lecture Notes in Computer Science, no.
  2683.\hskip 1em plus 0.5em minus 0.4em\relax Springer-Verlag, 2003.

\bibitem{cmucvsite}
{\tt http://www-2.cs.cmu.edu/\~{}cil/vision.html}.

\bibitem{Huynh:etal:iccv2003}
D.~Huynh, R.~Hartley, and A.~Heyden, ``Outlier correction in image sequences
  for the affine camera,'' in \emph{Proc. of IEEE Int. Conf. on Computer
  Vision}, Nice, France, 2003.

\bibitem{jacobs97}
D.~Jacobs, ``Linear fitting with missing data: Applications to
  structure-from-motion and to characterizing intensity images,'' in
  \emph{Proc. of IEEE Int. Conf. on Computer Vision and Pattern Recognition},
  1997.

\bibitem{maruyama99}
M.~Maruyama and S.~Kurumi, ``Bidirectional optimization for reconstructing 3{D}
  shape from an image sequence with missing data,'' in \emph{IEEE Proc. of Int.
  Conf. on Image Processing}, Kobe, Japan, 1999.

\bibitem{KN:McLachlanKrishnan}
G.~McLachlan and T.~Krishnan, \emph{The {EM} Algorithm and Extensions}.\hskip
  1em plus 0.5em minus 0.4em\relax New York: John Wiley \& Sons, 1997.

\bibitem{morita97}
T.~Morita and T.~Kanade, ``A sequential factorization method for recovering
  shape and motion from image streams,'' \emph{IEEE Trans. on Pattern Analysis
  and Machine Intelligence}, vol.~19, no.~8, 1997.

\bibitem{morris98}
D.~Morris and T.~Kanade, ``A unified factorization algorithm for points, line
  segments and planes with uncertainty models,'' in \emph{Proc. of IEEE Int.
  Conf. on Computer Vision}, 1998.

\bibitem{moses93}
Y.~Moses, ``Face recognition: generalization to novel images,'' Ph.D.
  dissertation, Weizmann Institute of Science, 1993.

\bibitem{poelman97}
C.~Poelman and T.~Kanade, ``A paraperspective factorization method for shape
  and motion recovery,'' \emph{IEEE Trans. on Pattern Analysis and Machine
  Intelligence}, vol.~19, no.~3, 1997.

\bibitem{quan96}
L.~Quan and T.~Kanade, ``A factorization method for affine structure from line
  correspondences,'' in \emph{Proc. of IEEE Int. Conf. on Computer Vision and
  Pattern Recognition}, San Francisco, CA, USA, 1996.

\bibitem{ripley96}
B.~Ripley, \emph{Pattern Recognition and Neural Networs}.\hskip 1em plus 0.5em
  minus 0.4em\relax Cambridge University Press, 1996.

\bibitem{roy-chowdhury04}
A.~Roy-Chowdhury, R.~Chellappa, and T.~Keaton, ``Wide baseline image
  registration with application to 3-{D} face modeling,'' \emph{IEEE Trans. on
  Multimedia}, vol.~6, no.~3, 2004.

\bibitem{shapiro95}
L.~Shapiro, \emph{Affine Analysis of Image Sequences}.\hskip 1em plus 0.5em
  minus 0.4em\relax Cambridge, UK: Cambridge University Press, 1995.

\bibitem{shashua92}
A.~Shashua, ``Geometry and photometry in 3{D} visual recognition,''
  Massachussets Institute of Technology, Cambridge MA, USA, MIT Technical
  Report 1401, 1992.

\bibitem{shum95}
H.~Shum, K.~Ikeuchi, and R.~Reddy, ``Principal component analysis with missing
  data and its applications to polyhedral object modeling,'' \emph{IEEE Trans.
  on Pattern Analysis and Machine Intelligence}, vol.~17, no.~9, 1995.

\bibitem{sturm96}
P.~Sturm and B.~Triggs, ``A factorization based algorithm for multi-image
  projective structure and motion,'' in \emph{Proc. of European Conf. on
  Computer Vision}, Cambridge, UK, 1996.

\bibitem{tomasi92}
C.~Tomasi and T.~Kanade, ``Shape and motion from image streams under
  orthography: a factorization method,'' \emph{Kluwer Int. Journal of Computer
  Vision}, vol.~9, no.~2, 1992.

\bibitem{ullman91}
S.~Ullman and R.~Basri, ``Recognition by linear combinations of models,''
  \emph{IEEE Trans. on Pattern Analysis and Machine Intelligence}, vol.~13,
  no.~10, 1991.

\bibitem{Vidal:Hartley:cvpr2004}
R.~Vidal and R.~Hartley, ``Motion segmentation with missing data using
  powerfactorization and {GPCA},'' in \emph{Proc. of IEEE Int. Conf. on
  Computer Vision and Pattern Recognition}, Washington DC, USA, 2004.

\bibitem{wiberg76}
T.~Wiberg, ``Computation of principal components when data are missing,'' in
  \emph{Proc. of Symposium on Computational Statistics}, Berlin, Germany, 1976.

\bibitem{manor99-1}
L.~Zelnik-Manor and M.~Irani, ``Multi-frame alignement of planes,'' in
  \emph{Proc. of IEEE Int. Conf. on Computer Vision and Pattern Recognition},
  Fort Collins, CO, USA, 1999.

\bibitem{manor99-2}
------, ``Multi-view subspace constraints on homographies,'' in \emph{IEEE Int.
  Conf. on Computer Vision}, Kerkyra, Greece, 1999.

\end{thebibliography}
\bibliographystyle{IEEEtranS}

\end{document}